\documentclass{article}

\PassOptionsToPackage{numbers, compress}{natbib}

 \usepackage[main, final]{neurips_2026}

\usepackage[utf8]{inputenc} 
\usepackage[T1]{fontenc}    
\usepackage{hyperref}       
\usepackage{url}            
\usepackage{booktabs}       
\usepackage{amsfonts}       
\usepackage{nicefrac}       
\usepackage{microtype}      
\usepackage{xcolor}         
\usepackage[dvipsnames]{xcolor}
\usepackage{amsmath}
\usepackage{amsthm}
\usepackage{amssymb}
\usepackage{bm}
\usepackage{graphicx}
\usepackage{cleveref}
\crefname{figure}{Figure}{Figures}
\Crefname{figure}{Figure}{Figures}
\usepackage{inconsolata}
\newtheorem{proposition}{Proposition}
\usepackage{natbib}
\usepackage{etoolbox}
\usepackage{array}
\usepackage{tcolorbox}
\tcbuselibrary{breakable}
\title{Jacobian Scopes: A Unified Geometric Framework for Token-Level LLM Attribution}

\author{%
  \textbf{Toni J.B. Liu}\textsuperscript{$\spadesuit$},\,
  \textbf{Baran Zadeoğlu}\textsuperscript{$\spadesuit$},\,
  \textbf{Gurbir Arora}\textsuperscript{$\spadesuit$},\,
  \\
  \textbf{Raphaël Sarfati}\textsuperscript{$\spadesuit\,\diamondsuit$},\,
  \textbf{Nicolas Boullé}\textsuperscript{$\heartsuit$},\,
  \textbf{Christopher J. Earls}\textsuperscript{$\spadesuit$}
\\
  \textsuperscript{$\spadesuit$}Cornell University, USA,\,
  \textsuperscript{$\heartsuit$}Imperial College London, UK, \,
  \textsuperscript{$\diamondsuit$}Goodfire AI, USA
\\
  \small{
    \textbf{Correspondence:} \href{mailto:jl3499@cornell.edu}{jl3499@cornell.edu}
  }
}

\begin{document}

\maketitle
\begin{abstract}
Large language models (LLMs) make next-token predictions based on clues present in their
context, such as semantic descriptions and in-context examples. Yet, elucidating which prior
tokens most strongly influence a given prediction remains challenging due to the proliferation of layers and attention heads in modern
architectures. We propose \emph{Jacobian Scopes},
a suite of gradient-based, token-level causal
attribution methods for interpreting LLM predictions. Grounded in perturbation theory and
information geometry, Jacobian Scopes quantify how input tokens influence various aspects
of a model’s prediction, such as specific logits, the full predictive distribution, and model
uncertainty (effective temperature). Through
case studies spanning instruction understanding, translation, and in-context learning (ICL),
we demonstrate how Jacobian Scopes reveal
implicit political biases, uncover word- and
phrase-level translation strategies, and shed
light on recently debated mechanisms underlying in-context time-series forecasting.
On token-ablation faithfulness benchmarks (LAMBADA and IWSLT2017) across six LLMs, the distribution-level Fisher and Temperature Scopes significantly outperform established gradient-based attribution methods in 10 of 12 model--dataset settings.
To facilitate exploration of Jacobian Scopes on custom text, we open-source our implementations and provide a cloud-hosted interactive demo at \url{https://huggingface.co/spaces/Typony/JacobianScopes}.
\end{abstract}

\newcommand{\Toni}[1]{{\color{blue}{TL: #1}}}

\section{Introduction}
Large Language Models (LLMs) exhibit surprising emergent abilities, 
such as multi-step reasoning \cite{huang-chang-2023-towards} and in-context learning (ICL) \cite{dong-etal-2024-survey,garg2023transformerslearnincontextcase}.
However, the mechanistic underpinning of these emergent abilities remains a deep and open question \cite{olsson2022context,reddy2023mechanisticbasisdatadependence,von2023transformers}. 
This has fueled a growing endeavor, across academia and industry, 
to interpret and explain LLMs' inference process during such emergent thinking modes, 
such as writing \cite{wan2023kellywarmpersonjoseph,guo2024biaslargelanguagemodels} and reasoning \cite{wei2022chain, wang2023understandingchainofthoughtpromptingempirical}. 

In this work, we present a suite of gradient-based causal attribution methods called \emph{Jacobian Scopes}.
Jacobian Scopes explain LLM predictions in terms of influential input tokens.
This is achieved by analyzing the Jacobian matrix, which captures the locally linearized relation between
the predictive distribution and input embeddings.

\begin{figure}[t]
  \centering
  \includegraphics[width=1\textwidth]{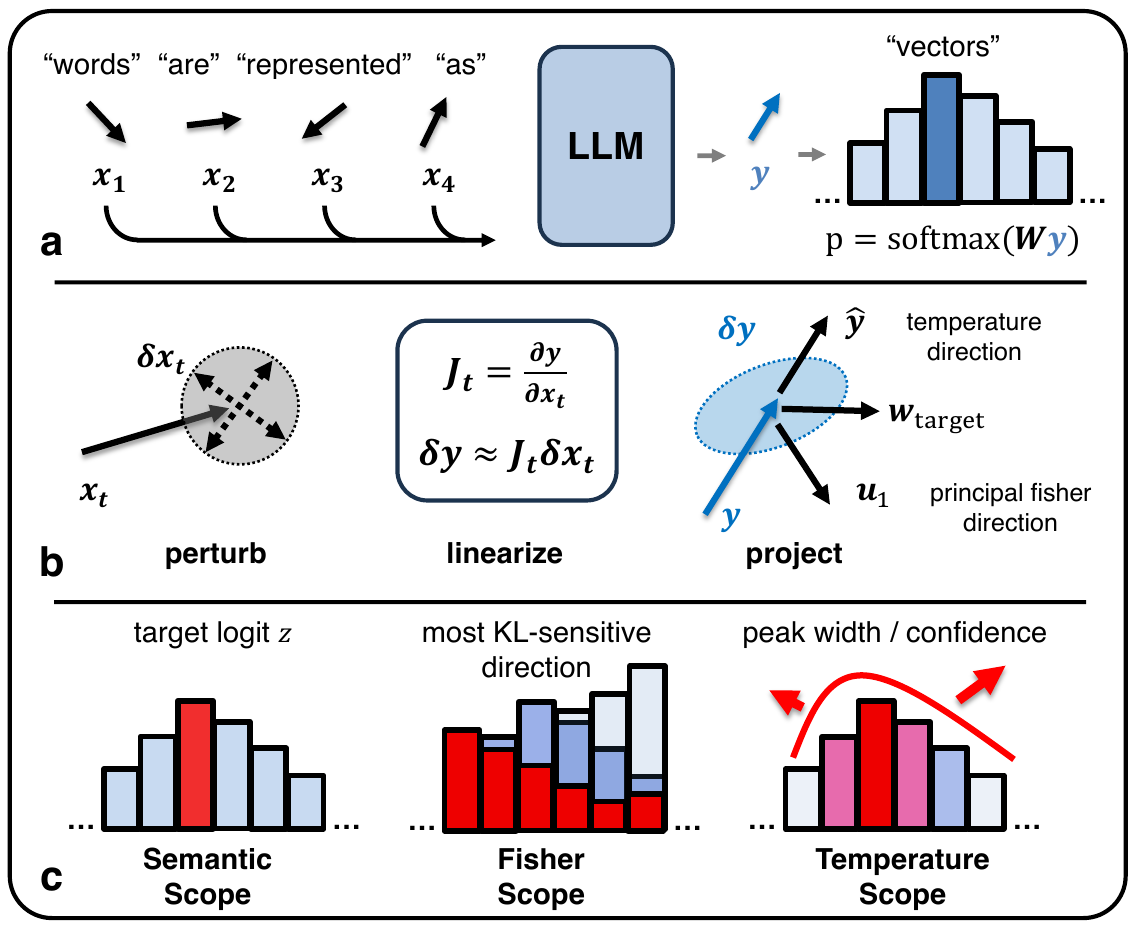}
  \caption{\textbf{Overview of Jacobian Scopes.}
  \textbf{(a)} An autoregressive LLM maps an input token sequence to a hidden state $\bm{y}$, which is decoded into a predictive distribution over the vocabulary.
  \textbf{(b)} A small perturbation to any input token is propagated to the output via the Jacobian $\bm{J}_t$, then projected onto a direction $\bm{v}$ in output space; the magnitude of this projection defines the token's influence score, computable in a single backward pass. The three Scopes correspond to $\bm{v} = \bm{w}_\mathrm{target}$ (target unembedding row), $\bm{v} = \bm{u}_1$ (principal Fisher direction), and $\bm{v} = \hat{\bm{y}}$ (normalized hidden state).
  \textbf{(c)} The three Jacobian Scope variants each target a different feature of the model's output: \textbf{Semantic Scope} asks which input tokens drive the predicted probability of a specific target word; \textbf{Fisher Scope} asks which tokens most change the overall shape of the predicted distribution; and \textbf{Temperature Scope} asks which tokens control the model's confidence, i.e., how sharply peaked the distribution is.}
  \label{fig:schematic}
\end{figure}

Jacobian Scopes come in three variants: \textbf{Semantic}, \textbf{Fisher}, and \textbf{Temperature Scopes}, each corresponding to a different \emph{explanandum}, or attribution objective: the logit of a specific token, the full predictive distribution, and the model's uncertainty (effective temperature). 
An overview of the framework is illustrated in \cref{fig:schematic}.

We showcase the efficacy of our proposed Jacobian Scopes through a range of case studies,
including natural-language instructions such as system prompts and translation,
as well as ICL tasks such as 
time-series forecasting  \cite{gruver2024largelanguagemodelszeroshot,mirchandani2023largelanguagemodelsgeneral}.
We further evaluate attribution faithfulness quantitatively on two benchmark datasets (LAMBADA \cite{paperno-etal-2016-lambada} and IWSLT2017 \cite{cettolo-etal-2017-overview}) 
across three suites of leading LLMs: LLaMA-3.2 \cite{llama3-2024}, Qwen2.5 \cite{qwen2025qwen25} and Gemma-3 \cite{gemmateam2025gemma3technicalreport}.

\section{Related work}
\label{sec:related_work}
Gradient-based attribution methods were first systematically developed for image classifiers.
\citet{simonyan2013deep} introduced class saliency maps by computing the
gradient of the predicted class score with respect to the input pixels,
identifying which image regions drive a given classification decision in a
single backward pass.
Building on this, \citet{shrikumar2017learning} proposed DeepLIFT, which
propagates neuron contribution scores by comparing each activation to a
reference state; gradient $\times$ input --- the elementwise product of the
input embedding and its gradient with respect to the target logit --- arises
as a simpler special case and serves as a baseline in our evaluations.
Grad-CAM \citep{Ramprasaath-Grad-Cam-2017} extends gradient saliency to produce spatially coarser but more semantically coherent maps by pooling gradients across feature channels.

A key challenge with plain gradient saliency is that the resulting maps are
often noisy and fail to sharply localize the relevant input features.
SmoothGrad \citep{smilkov2017smoothgradremovingnoiseadding} addresses this for
vision models by averaging sensitivity maps over many copies of the input
perturbed with Gaussian noise.
Integrated Gradients \citep{Sundararajan-Axiomatic-Attribution-DNN-2017}
instead eliminates path-dependence by integrating the gradient along a linear
interpolation from a null baseline to the actual input, satisfying axiomatic
desiderata such as Sensitivity and Implementation Invariance.
\citet{sanyal2021discretizedintegratedgradientsexplaining}
subsequently adapted this path integration to the discrete embedding space of
language models.
Outside the gradient paradigm, SHAP \citep{lundberg2017unified} provides model-agnostic attribution scores grounded in Shapley values from cooperative game theory, offering strong axiomatic guarantees but at combinatorial cost that must be approximated in practice.

A complementary line of work interprets transformers through the lens of attention weights.
\citet{Clark-Berts-Attention-2019} visualize BERT's attention patterns and identify heads that track syntactic and coreference relations.
\citet{abnar-zuidema-2020-quantifying} extend this with attention rollout and attention flow, which propagate attention weights across layers to produce more faithful input-attribution scores.
While intuitive and computationally cheap, these attention-based scores do not directly measure the causal influence of input tokens on model outputs, and can be misleading as attribution methods \citep{heimersheim2024useinterpretactivationpatching}.
\citet{ferrando-etal-2022-towards} address this for neural machine translation by proposing ALTI (Aggregated LaTent Interaction), a gradient-based method that attributes each generated target token to specific source tokens; our Fisher Scope case study on translation (\cref{fig:translating Dante}) can be viewed as a distributional generalization of this attribution objective.

More recently, mechanistic interpretability methods have sought to go beyond input-output sensitivity and trace computations through specific model circuits.
Activation patching \citep{heimersheim2024useinterpretactivationpatching, pmlr-v235-ghandeharioun24a} localizes responsible components by intervening on intermediate representations, while circuit tracing \citep{ameisen2025circuit} reconstructs full computational graphs.
Sparse autoencoder (SAE) probes \citep{lieberum2024gemmascopeopensparse} decompose residual stream activations into interpretable features.
These methods offer fine-grained mechanistic insights, but are substantially more expensive and architecture-specific than gradient-based approaches.

All gradient-based methods described above attribute a \emph{scalar model output}---a single class logit or token log-likelihood---and are therefore conceptually analogous to \emph{Semantic Scope}.
Yet LLMs produce a full distribution over the vocabulary, and, to the best of our knowledge, no prior method exploits this richer structure.

\section{Methodology}
\label{sec:methodology}

\textbf{Notation.} We use bold lowercase letters (e.g.\ $\bm{x}$) for vectors and bold uppercase letters (e.g.\ $\bm{W}$) for matrices. 

Let $\mathcal H := \mathbb R^{d_{\text{model}}}$ denote the model’s hidden space. We represent a length-$T$ input sequence as
$\bm X_{1:T} := (\bm{x}_1,\ldots, \bm{x}_t, \ldots, \bm{x}_T)$, where $\bm{x}_t \in \mathcal H$.
We view an autoregressive LLM as a function that maps a sequence of inputs
to the final-layer, post-layer-norm hidden state at the leading position:
\[
\bm y := f(\bm{x}_1,\ldots, \bm{x}_t, \ldots, \bm{x}_T) \in \mathcal H.
\]
The corresponding logits~$\bm{z}$ and predictive distribution~$\bm{p}$ are given by $\bm{z} := \bm{W}\bm{y}$ and $\bm{p}(\cdot \mid \bm{X}_{1:T}) := \mathrm{softmax}(\bm{z}) \in \mathbb{R}^{|\mathcal{V}|}$, where \(\bm{W} \in \mathbb{R}^{|\mathcal{V}| \times d_{\mathrm{model}}}\) denotes the unembedding matrix, 
and \(\mathcal{V}\) is the vocabulary set of size $|\mathcal{V}|$.

\noindent
\textbf{Definition of influence score.} 
We are interested in quantifying how perturbations to an input token embedding \(\bm{x}_t\) affect the leading hidden state \(\bm{y}\).
We define the input-to-output Jacobian at position \(t\) as
\[
\bm{J}_t := \frac{\partial \bm{y}}{\partial \bm{x}_t}\;\in\;\mathbb{R}^{d_{\mathrm{model}}\times d_{\mathrm{model}}}.
\]
Forming the full Jacobian \(\bm{J}_t\) requires \(d_{\mathrm{model}}\) backward passes, which can be expensive for modern LLMs.
Fortunately, in practice we find it sufficient to compute the vector-Jacobian product, $\bm{v}^\intercal \bm{J}_t$, where \(\bm{v} \in \mathcal{H}\) captures a property of interest in the predictive distribution, such as confidence (effective inverse temperature) or the embedding of a target token.

Geometrically, $\lVert \bm{v}^\intercal \bm{J}_t \rVert_2$ measures the largest displacement 
along the direction $\bm{v}$ 
that could be induced by an $\varepsilon$-norm perturbation to \(\bm{x}_t\) (proven in Appendix \ref{app:epsilon_perturbation}).
We thus define the \emph{influence} of token \(\bm{x}_t\) on the leading output \(\bm{y}\) as
\begin{equation}
\mathrm{Influence}_t := \left\lVert \bm{v}^\intercal \bm{J}_t \right\rVert_2.
\label{def:influence}
\end{equation}

\noindent
\textbf{Influence computation via auto-diff.} 
For the leading prediction \(\bm{y}\), 
the influence scores at all input positions \(t \le T\) can be computed with a single backward pass by auto-differentiating the customized scalar loss
$\mathcal{L}(\bm{X}_{1:T}) := \bm{v}^\intercal \bm{y}$:
\begin{equation}\label{influence via auto-diff}
  \left\lVert \frac{\partial \mathcal{L}(\bm{X}_{1:T}) }{\partial \bm{x}_t} \right\rVert_2
    = \left\|\bm{v}^\intercal \frac{\partial \bm{y}}{\partial \bm{x}_t}\right\|_2= \mathrm{Influence}_t.
\end{equation}
In \cref{sec:applications}, we present three specific choices for the direction of interest $\bm{v}$, leading to \emph{Semantic Scope}, \emph{Fisher Scope} and \emph{Temperature Scope}; see \cref{fig:schematic} for an overview. 

\section{Example applications}
\label{sec:applications}
This section discusses three formulations of Jacobian Scopes, each motivated by a specific attribution objective.
For consistency, all demonstrations in this section are conducted on LLaMA-3.2 1B.

\subsection{Semantic Scope}
\label{sec:Semantic Scope}

\textit{Semantic Scope} traces the input tokens that most prominently contribute to the predicted probability of a target token. Here, the direction of interest is $\bm{v}=\bm{w}_\mathrm{target}$, which is the row of the unembedding matrix $\bm{W}$ corresponding to the target token. The customized scalar loss becomes
\begin{equation}
  \mathcal{L}_\mathrm{semantic}:= z_\mathrm{target} = \bm{w}_\mathrm{target}^\intercal \bm{y},
\end{equation}
which is exactly the predicted logit of the target token. 
Differentiating this loss using \cref{influence via auto-diff} yields the Semantic Scope influence score
\begin{equation}
  \mathrm{Influence}_t^{\mathrm{Sem}} := \left\|\bm{w}_\mathrm{target}^\intercal \bm{J}_t\right\|_{2}.
\end{equation}

\begin{figure}[h]
\centering
  \includegraphics[width=0.8\columnwidth]{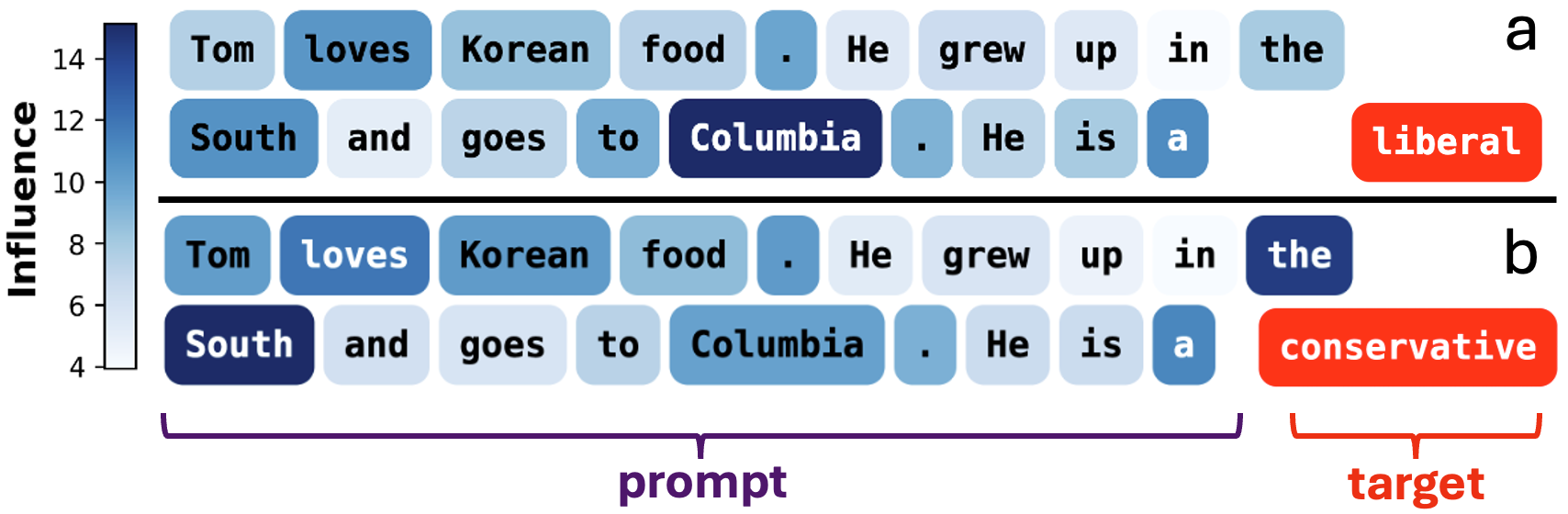}
  \caption{Semantic Scope reveals the potential for political bias that may be present in models such as LLaMA-3.2. The prediction for the subject being a ``liberal'' is attributed to the input token ``Columbia'', while the prediction for ``conservative'' is attributed to the token pair ``the South''.
  }
  \label{fig:causal attribution 2}
\end{figure}

Intuitively, Semantic Scope quantifies how sensitive the target logit is to changes in specific input tokens. 
\Cref{fig:causal attribution 2} demonstrates how Semantic Scope may be employed to visualize implicit political bias in LLMs (see also \cref{fig:causal attribution 1} in the appendix for a system-prompt example).

\subsection{Fisher Scope}
\label{sec:fisher_scope}
\begin{figure}[ht]
  \includegraphics[width=1\columnwidth]{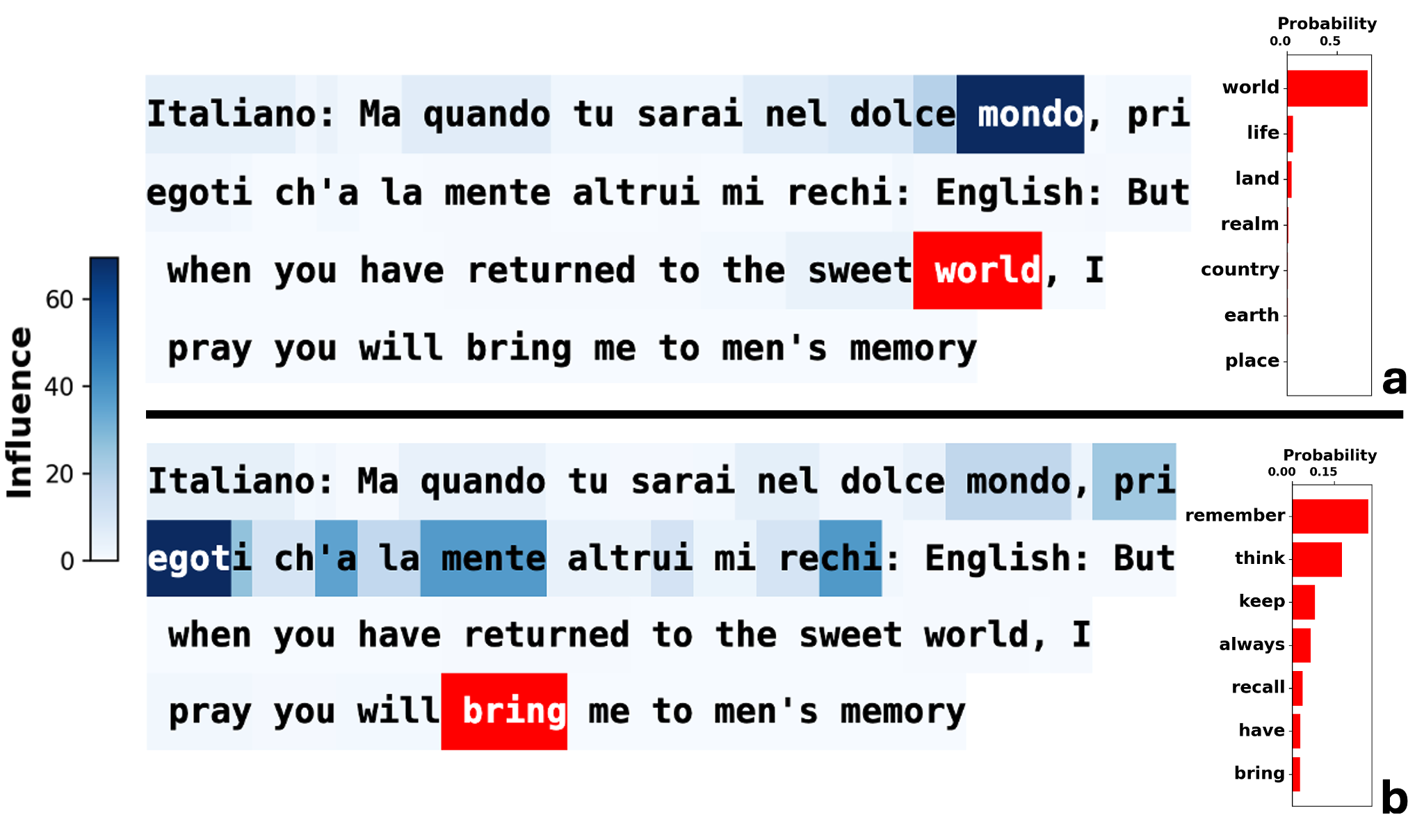}
  \caption{Fisher Scope visualizes LLaMA-3.2 3B's strategy for translating Dante's Inferno \cite{hollander2002inferno}.
  Left: Magnitude of influence from each source word (blue) on the predicted distribution at each target position (red).
  Right: Probability distribution of the top 7 words predicted at each target position.
  (a) LLaMA assigns high probability to ``world'', the unambiguous literal translation of ``mondo''.
  (b) When translating the phrase ``priegoti ch'a la mente altrui mi rechi'' (literally: ``I pray thee to bring me to another's mind''), LLaMA's predicted distribution is highly uncertain, with each top token hinting at a different phrasing strategy.}
  \label{fig:translating Dante}
\end{figure}

Semantic Scope attributes the past tokens that affect a specific output logit, 
but it does not capture their influence on the \emph{entire} predictive distribution $\bm{p}(\cdot \mid \bm{X}_{1:T})$. 
The latter motivates Fisher Scope, which is particularly suited for attribution tasks where the correct next-word prediction is non-unique, 
such as translation (see \cref{fig:translating Dante}(b)).

Fisher Scope is inspired by information geometry \cite{fisher1922mathematical}, which studies how infinitesimal changes in model parameters affect the predictive distributions \cite{bishop2006pattern}.
In our setting, the hidden state $\bm{y}$ parameterizes the family $\bm{p}(\bm{y}) = \mathrm{softmax}(\bm{W}\bm{y})$, and the Fisher information matrix (FIM) $\bm{F}$, derived in \cref{app:kl_second_order}, measures how sensitively $\bm{p}$ responds to perturbations in $\bm{y}$:
\begin{equation}
\bm{F} := \bm{W}^\intercal \big(\mathrm{diag}(\bm{p}) - \bm{pp}^\intercal\big) \bm{W}.
\label{eq:fisher_metric}
\end{equation}
Intuitively, the spectrum of 
$\bm{F}$ reveals how sensitively a perturbation along each eigendirection can change the 
predictive distribution. 
Formally, the eigen-decomposition
$\bm{F} = \bm{U}\bm{\Lambda}\bm{U}^\intercal$
assigns to each direction $\bm{u}_i$ a scalar sensitivity $\lambda_i$, quantifying 
how strongly a perturbation along $\bm{u}_i$ changes 
$\bm{p}(\cdot \mid \bm{X}_{1:T})$, as measured by KL divergence 
\cite{amari2000methods}.

We call the leading eigenvector 
$\bm{u}_1$ the \emph{principal Fisher direction}: it is the single 
direction in hidden space $\mathcal{H}$ along which a perturbation to $\bm{y}$ most 
sensitively changes $\bm{p}(\cdot \mid \bm{X}_{1:T})$.
The Fisher Scope influence score is then defined as the projection of 
the Jacobian onto this direction:
\begin{equation}
    \mathrm{Influence}_t^{\mathrm{Fisher}} := \|\bm{u}_1^\intercal \bm{J}_t\|_2.
\label{eq: fisher influence}\end{equation}
Why project onto $\bm{u}_1$ rather than attribute the change in the distribution directly?
Below we motivate this design choice by analyzing, to second order, the information geometry induced by the softmax.

\noindent\textbf{Second-order information at first-order cost.}
The natural measure of how a perturbation $\bm{\delta x}_t$ to token $t$ changes the prediction as a whole is the KL divergence between the original and perturbed predictive distributions,
$\mathrm{KL}_t(\bm{\delta x}_t) := \mathrm{KL}\bigl(\bm{p}(\cdot\mid\bm{x}_t)\,\|\,\bm{p}(\cdot\mid\bm{x}_t+\bm{\delta x}_t)\bigr)$.
Because $\mathrm{KL}_t$ attains its minimum of zero at $\bm{\delta x}_t=\bm{0}$, its gradient vanishes there, so any first-order attribution of distributional change is identically zero.
The leading behavior is instead quadratic (\cref{app:kl_second_order}):
\begin{equation}
\mathrm{KL}_t(\bm{\delta x}_t)
= \tfrac{1}{2}\,\bm{\delta x}_t^\intercal\,\bm{F}_t\,\bm{\delta x}_t + o\bigl(\|\bm{\delta x}_t\|^2\bigr),
\qquad
\bm{F}_t := \bm{J}_t^\intercal \bm{F}\,\bm{J}_t ,
\label{eq:kl_main}
\end{equation}
where $\bm{F}_t$, the Fisher information pulled back to the embedding of token $t$, is the \emph{exact} Hessian of $\mathrm{KL}_t$: the curvature of the network itself drops out because the gradient of $\mathrm{KL}_t$ vanishes at the unperturbed input.
Averaging over small isotropic perturbations $\bm{\delta x}_t = \varepsilon\bm{\xi}$, with $\varepsilon \ll 1$ and $\bm{\xi}$ uniform on the unit sphere, gives, to leading order in $\varepsilon$ and up to a position-independent constant (\cref{app:fisher_total_information}),
\begin{equation}
\mathbb{E}_{\bm{\xi}}\bigl[\mathrm{KL}_t(\varepsilon\bm{\xi})\bigr]
\propto \mathrm{tr}(\bm{F}_t)
= \sum_{i=1}^{d_{\mathrm{model}}} \lambda_i \,\|\bm{u}_i^\intercal\bm{J}_t\|_2^2 ,
\label{eq:total_info_main}
\end{equation}
so the \emph{total information} $\mathrm{tr}(\bm{F}_t)$ decomposes along the Fisher eigendirections, and Fisher Scope retains its leading term, $\lambda_1\bigl(\mathrm{Influence}^{\mathrm{Fisher}}_t\bigr)^2$.
Fisher Scope is therefore second-order in the induced distribution shift.
Yet because $\bm{F}$ is available in closed form from the logits (\cref{eq:fisher_metric}), Fisher Scope requires no additional passes through the network: a single backward pass suffices, the same cost as a first-order gradient.
\Cref{fig:fisher_rank_aopc} shows that this rank-1 truncation loses little faithfulness in practice.

\Cref{fig:translating Dante} demonstrates how Fisher Scope reveals two non-exclusive translation strategies employed by LLaMA-3.2.
(a) When the source text has precise word-level counterparts, the LLM opts for word-to-word translation, with each predicted token attributed to its lexical counterpart.
(b) When word-to-word translation is not possible due to mismatched syntactic structures, the LLM works at the level of phrases, with each predicted position attending to multiple source tokens.
Fisher Scope is particularly well-suited to case (b): because the predicted distribution becomes highly uncertain across many plausible phrasings, a target-logit method like Semantic Scope would only attribute the influence on one arbitrarily chosen token, whereas Fisher Scope captures the full distributional uncertainty and attributes it to the relevant source phrase.
We refer the reader to \cref{sec: additional experiments} for a simpler word-level translation example.

\subsection{Temperature Scope}
\label{sec:Temperature Scope}

\begin{figure}[ht]
\centering
  \includegraphics[width=0.9\columnwidth]{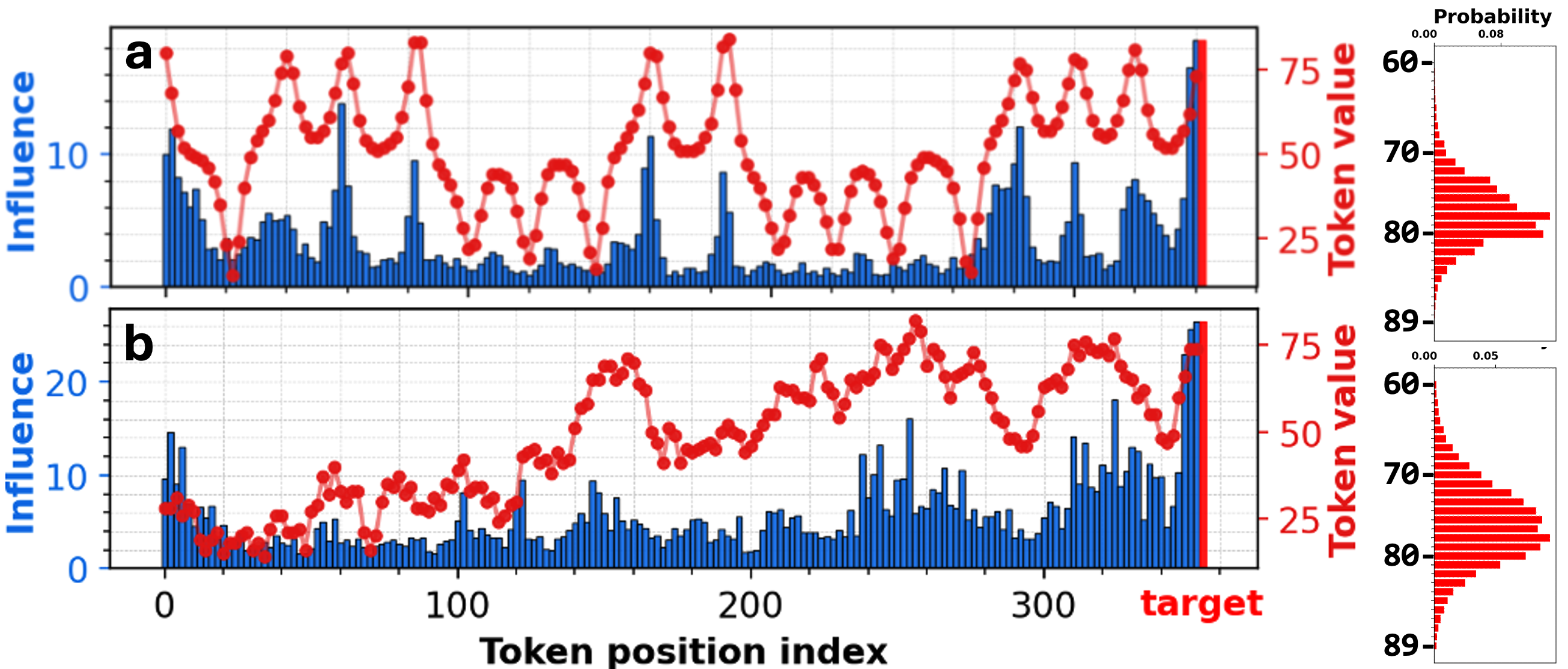}
  \caption{Left: Temperature Scope reveals different strategies employed by LLaMA-3.2 during time-series forecasting:
  (a) When continuing the dynamics of a partially observed Lorenz system, LLaMA attends to regions in the input history that exhibit patterns similar to those near the cutoff. (b) When extrapolating unstructured Brownian motion, LLaMA shows diminished influence of early context.
  Right: Predicted probability distribution over the top 30 most probable tokens.}
  \label{fig:lorenz brownian}
\end{figure}

While semantic next-word predictions span a large and heterogeneous vocabulary, 
structured ICL tasks, such as time-series prediction \cite{gruver2024largelanguagemodelszeroshot}, often exhibit structural regularities: 
the predicted distributions form near-Gaussian peaks over a small subset of numerical tokens (see \cref{fig:lorenz brownian,fig:JS_lorenz} as well as \cite{liu2025density}).
Temperature Scope identifies which input tokens control the width of this Gaussian peak, and hence the model's forecasting uncertainty.

To define Temperature Scope, we first decompose the hidden state $\bm{y}$ into its norm and direction, $\bm{y} = \|\bm{y}\|_2 \hat{\bm{y}}$, 
so that the logit vector can be written as 
\[
  \bm{z} = \bm{W}\bm{y} = \|\bm{y}\|_2\,(\bm{W}\hat{\bm{y}}) = \beta_\mathrm{eff}\,\hat{\bm{z}},
\]
where $\beta_\mathrm{eff} := \|\bm{y}\|_2$ is the effective inverse temperature and $\hat{\bm{z}}$ the normalized logits.

The key theoretical justification is that $\beta_\mathrm{eff}$ directly controls predictive spread.
Treating token values $v$ as continuous, let $p(v) \propto e^{\,\beta_\mathrm{eff}\,\hat{z}(v)}$ be the predicted distribution.

\begin{proposition}[Effective temperature as inverse variance]
\label{prop:beta_variance}
$p(v)$ is approximately Gaussian if and only if $\hat{z}(v)$ is quadratic in the neighborhood of the peak.
Writing $\hat{z}(v) = -b(v-\mu)^2 + c$ with $b>0$, the predictive variance satisfies
\begin{equation}
    \sigma^2 = \frac{1}{2\,\beta_\mathrm{eff}\, b},
    \label{eq:beta_variance}
\end{equation}
so $\beta_\mathrm{eff} \propto \sigma^{-2}$, with $b$ a model-dependent constant.
\end{proposition}

The proof follows directly from matching the Gaussian kernel to the Boltzmann form; see \cref{app:beta_variance}.
\Cref{prop:beta_variance} is intended for the numerical ICL setting: when the predictive distribution over numerical tokens is approximately Gaussian (as in \cref{fig:lorenz brownian}), attributing $\beta_\mathrm{eff}$ is equivalent to attributing predictive variance.
This motivates using $\mathcal{L}_\mathrm{temperature} = \beta_\mathrm{eff}$ as the customized loss, whose auto-differentiation yields the influence score:
\begin{equation}
  \mathrm{Influence}_t^{\mathrm{Temp}} :=\left\| \hat{\bm{y}}^\intercal \bm{J}_t \right\|_2.
  \label{eq:temp_influence}
\end{equation}

\noindent\textbf{Entropy interpretation for general vocabularies.}
For unstructured natural-language prediction, a continuous-token approximation is undefined, so \cref{prop:beta_variance} does not apply directly.
Temperature Scope is nonetheless competitive on LAMBADA and IWSLT2017 (\cref{tab:aupc}).
A simple explanation is that, for any discrete softmax with fixed logit \emph{shape}, attributing inverse temperature is equivalent to attributing Shannon entropy---and hence predictive uncertainty.

Write $p(i;\beta)\propto e^{\beta\hat{z}(i)}$ with normalized logits $\hat{\bm{z}}$ held fixed, and let
$V(\beta)=\mathrm{Var}_{i\sim p(\cdot;\beta)}[\hat{z}(i)]$.
Differentiating the Shannon entropy $H(\beta)=-\sum_i p(i;\beta)\log p(i;\beta)$ gives
\begin{equation}
  \frac{dH}{d\beta}
  = -\beta\, V(\beta).
  \label{eq:dH_dbeta}
\end{equation}
Whenever $\beta>0$ and $\hat{\bm{z}}$ is non-constant, $V(\beta)>0$, so $H$ is strictly decreasing in $\beta$:
$H\to\log|\mathcal{V}|$ as $\beta\to 0$ and $H\to 0$ as $\beta\to\infty$.
Thus $\beta\mapsto H$ is invertible, and attributing $\beta_\mathrm{eff}$ is equivalent to attributing $H$.

The correspondence extends to token-level scores.
Temperature Scope computes the radial sensitivity of $\beta_\mathrm{eff}$ to each input while holding $\hat{\bm{z}}$ fixed, so
\begin{equation}
  \frac{\partial H}{\partial \bm{x}_t}
  = -\beta\, V(\beta)\,
  \frac{\partial \beta_\mathrm{eff}}{\partial \bm{x}_t}
  \qquad(\hat{\bm{z}}\ \text{fixed}).
  \label{eq:entropy_attribution}
\end{equation}
The prefactor $-\beta V(\beta)$ is shared by every token, so Temperature Scope and entropy attribution induce the same gradient-norm ranking.
In short, there are two complementary readings of the same radial Jacobian projection: on numerical ICL, Temperature Scope attributes predictive \emph{variance}; on general text, it attributes predictive \emph{entropy}.

To probe how an LLM reasons over its inputs during numerical ICL tasks,
we follow \citet{gruver2024largelanguagemodelszeroshot} and tokenize time-series data as alternating commas and 2-digit numbers,\footnote{In LLaMA-3.2, there is one token for each 2-digit number from 10 to 99.}
yielding purely numerical prompts such as $29,30,45,\ldots$.
Temperature Scope then attributes the confidence of the predicted next-state distribution to the inputs.

As shown in \cref{fig:lorenz brownian}, 
when extrapolating chaotic time-series data with recurring but non-identical motifs,
LLaMA exhibits consistent attribution patterns: it attends to segments in history that match the local pattern near the cutoff point.
This attribution signature directly supports the recent conjecture that LLMs might parrot the best-matching pattern
in history via nearest-neighbor search in delay-embedding space \cite{zhang2025contextparrotingsimpletoughtobeat}.

On the other hand, when faced with stochastic systems with no recurring motifs, 
such as Brownian motion, the LLM employs a different strategy: it preferentially attends to later parts of the context. This strategy may explain the observation \cite{liu-etal-2024-llms-learn} that the ICL loss curve plateaus early for unbounded stochastic systems such as Brownian motion and geometric Brownian motion: as such systems wander out of distribution, the LLM ceases to recognize earlier history as relevant, and gradually forgets earlier statistics. 
We investigate this hypothesis further in \cref{Additional experiments on Dynamical Systems}.

\Cref{app:temperature_optimal_num} further shows that Temperature Scope produces more interpretable attributions of in-context forecasting behavior than Semantic and Fisher Scopes.

\section{Quantitative Evaluations}
\label{sec:evaluation}

\begin{table}[!t]
  \centering
  \small
  \caption{AOPC (higher magnitude is better) on LAMBADA (LMB) and IWSLT2017 DE$\to$EN (IWS)
  for all six models, averaged over ablation fractions
  $k \in \{5\%, 10\%, 20\%\}$ via the trapezoidal rule over 1000 test passages.}
  \renewcommand{\arraystretch}{1.3}
  \setlength{\tabcolsep}{3pt}
  \medmuskip=1mu
  \resizebox{\textwidth}{!}{%
  \begin{tabular}{lcccccc}
  \multicolumn{7}{c}{\textbf{1B/1.5B-scale}} \\
  \hline
    & \multicolumn{2}{c}{\textbf{LLaMA-3.2 1B}} & \multicolumn{2}{c}{\textbf{Qwen2.5-1.5B}} & \multicolumn{2}{c}{\textbf{Gemma-3 1B}} \\
  \cline{2-3} \cline{4-5} \cline{6-7}
  \textbf{Method} & \textbf{LMB} & \textbf{IWS} & \textbf{LMB} & \textbf{IWS} & \textbf{LMB} & \textbf{IWS} \\
  \hline
  Random              & $-0.26 \pm 0.01$ & $-0.30 \pm 0.01$ & $-0.35 \pm 0.01$ & $-0.25 \pm 0.01$ & $-0.42 \pm 0.01$ & $-0.41 \pm 0.01$ \\
  Integrated Grads    & $-1.19 \pm 0.01$ & $-1.04 \pm 0.01$ & $-1.71 \pm 0.01$ & $-1.36 \pm 0.01$ & $-0.69 \pm 0.01$ & $-0.87 \pm 0.01$ \\
  Input $\times$ Grad & $-1.28 \pm 0.01$ & $-1.04 \pm 0.01$ & $-1.72 \pm 0.01$ & $-1.35 \pm 0.01$ & $-1.56 \pm 0.01$ & $-1.36 \pm 0.01$ \\
  \hline
  Semantic Scope      & $-1.30 \pm 0.01$ & $-1.06 \pm 0.01$ & $-1.68 \pm 0.01$ & $-1.24 \pm 0.01$ & $-1.64 \pm 0.01$ & $-1.30 \pm 0.01$ \\
  Temperature Scope   & $\mathbf{-1.32 \pm 0.01}$ & $\mathbf{-1.09 \pm 0.01}$ & $\mathbf{-1.86 \pm 0.01}$ & $\mathbf{-1.38 \pm 0.01}$ & $-1.66 \pm 0.01$ & $-1.29 \pm 0.01$ \\
  Fisher Scope        & $\mathbf{-1.32 \pm 0.01}$ & $-1.08 \pm 0.01$ & $-1.77 \pm 0.01$ & $\mathbf{-1.38 \pm 0.01}$ & $\mathbf{-1.67 \pm 0.01}$ & $\mathbf{-1.39 \pm 0.01}$ \\
  \hline
  \\[2pt]
  \multicolumn{7}{c}{\textbf{3B/4B-scale}} \\
  \hline
   & \multicolumn{2}{c}{\textbf{LLaMA-3.2 3B}} & \multicolumn{2}{c}{\textbf{Qwen2.5 3B}} & \multicolumn{2}{c}{\textbf{Gemma-3 4B}} \\
  \cline{2-3} \cline{4-5} \cline{6-7}
  \textbf{Method} & \textbf{LMB} & \textbf{IWS} & \textbf{LMB} & \textbf{IWS} & \textbf{LMB} & \textbf{IWS} \\
  \hline
  Random              & $-0.23 \pm 0.01$ & $-0.19 \pm 0.01$ & $-0.29 \pm 0.01$ & $-0.20 \pm 0.01$ & $-0.29 \pm 0.01$ & $-0.18 \pm 0.01$ \\
  Integrated Grads    & $-0.69 \pm 0.01$ & $-0.61 \pm 0.01$ & $-1.39 \pm 0.01$ & $-1.12 \pm 0.01$ & $-1.50 \pm 0.02$ & $\mathbf{-1.34 \pm 0.01}$ \\
  Input $\times$ Grad & $-1.12 \pm 0.01$ & $-0.77 \pm 0.01$ & $-1.39 \pm 0.01$ & $-1.11 \pm 0.01$ & $-1.70 \pm 0.02$ & $-1.17 \pm 0.01$ \\
  \hline
  Semantic Scope      & $-1.16 \pm 0.01$ & $-0.78 \pm 0.01$ & $-1.34 \pm 0.01$ & $-1.01 \pm 0.02$ & $-1.78 \pm 0.02$ & $-1.06 \pm 0.01$ \\
  Temperature Scope   & $\mathbf{-1.17 \pm 0.01}$ & $-0.76 \pm 0.01$ & $\mathbf{-1.55 \pm 0.01}$ & $\mathbf{-1.20 \pm 0.01}$ & $-1.78 \pm 0.02$ & $-0.99 \pm 0.01$ \\
  Fisher Scope        & $\mathbf{-1.17 \pm 0.01}$ & $\mathbf{-0.80 \pm 0.01}$ & $-1.41 \pm 0.01$ & $-1.15 \pm 0.01$ & $\mathbf{-1.81 \pm 0.01}$ & $-1.20 \pm 0.01$ \\
  \hline  
  \end{tabular}}
  \label{tab:aupc}
\end{table}

To complement the qualitative case studies in \cref{sec:applications}, 
we provide quantitative evaluations against established attribution baselines across two benchmarks.

\noindent\textbf{Datasets.}
We evaluate on two tasks that probe complementary aspects of token-level attribution.
\begin{itemize}
    \item \textbf{LAMBADA} \cite{paperno-etal-2016-lambada} is a language modeling benchmark designed to test long-range context comprehension. 
    The dataset is constructed such that a human subject would need to locate clues scattered across many sentences in order to correctly predict the final word of each passage.
    \item \textbf{IWSLT2017 DE$\to$EN} \cite{cettolo-etal-2017-overview} is a machine translation dataset that contains passages in German followed by English translations.

\end{itemize}

Together, these datasets test whether Jacobian Scopes produce faithful attributions across both monolingual and cross-lingual generation.
We compare against random ablation, Integrated Gradients \cite{Sundararajan-Axiomatic-Attribution-DNN-2017}, and Input $\times$ Gradient \cite{shrikumar2017learning,simonyan2013deep}.
Full experimental details are provided in \cref{sec:benchmark_details}; PAD-token ablation and the full paired-test table appear in \cref{sec:pad_ablation,sec:paired_full}.
Snapshots of each dataset and their respective Fisher Scope attributions are shown in \cref{fig:lambada_iwslt_examples} in the appendix, giving a sense of the typical prompt structure and attribution patterns.

\noindent\textbf{Evaluation metric.}
We evaluate using the \textbf{Area Over the Perturbation Curve (AOPC)} \cite{samek2016evaluating}.
AOPC is a counterfactual, interventional test: the top-$k$\% most influential tokens are ablated (replaced with zero vectors), and the resulting drop in the ground-truth target token's log-probability is recorded.

If an attribution method correctly identifies the tokens that the model relies on, removing them should cause a large, predictable drop in model confidence.
This makes AOPC a particularly demanding benchmark for gradient-based methods such as Jacobian Scopes, which rely only on local derivatives of the network: performing well on this interventional metric implies that local sensitivity is a reliable proxy for genuine causal relevance.

For robustness and consistency with the literature \cite{samek2016evaluating, deyoung-etal-2020-eraser}, we zero out the most influential tokens at three distinct ablation rates, $k \in \{5\%, 10\%, 20\%\}$. The final AOPC is then aggregated via the trapezoidal rule:
\begin{align}
    \mathrm{AOPC} = \sum_{i} \tfrac{1}{2}(k_{i+1} - k_i)
    \left(\Delta\log p_{k_i} + \Delta\log p_{k_{i+1}}\right),
\end{align}
where $\Delta \log p_k$ is the drop in log-probability when the top-$k$\% tokens are ablated.
A more negative AOPC indicates higher attribution faithfulness.

\noindent\textbf{Results.}
\Cref{tab:aupc} reports AOPC across all six models at both 3B/4B and 1B/1.5B scales.
Jacobian Scopes outperform random ablation and match or exceed the strongest baseline in most settings; the main exception is Gemma-3 4B on IWSLT2017, where Integrated Gradients (20-point path integral) attains the best mean AOPC.
To assess whether these gaps are reliable at the passage level, we compare, for each model--dataset cell, the mean over passages of
$\mathrm{AOPC}_i(\mathrm{Scope})-\mathrm{AOPC}_i(\mathrm{baseline})$,
where the Scope and non-Scope baseline are the best methods in that cell of \cref{tab:aupc}.
Negative values favor the Scope.
The Scope wins significantly in \textbf{10 of 12} combinations, with one inconclusive comparison and one significant baseline win (Gemma-3 4B IWSLT2017); see \cref{tab:paired_aopc_full} for method pairs, confidence intervals, and $p$-values.
As a robustness check, we also rerun AOPC with length-preserving PAD-token ablation in \cref{sec:pad_ablation}.
We discuss this model- and task-specific pattern in \cref{sec:discussion}.

These results reveal a perhaps surprising finding: Fisher and Temperature Scopes, which are entirely \emph{target-blind}, match or outperform the target-specific methods (Semantic Scope, Input $\times$ Gradient, Integrated Gradients) under AOPC in most settings, despite AOPC being an intrinsically target-specific metric that measures the drop in the ground-truth token's log-probability upon ablation.

One might expect target-specific methods to have an inherent advantage here.
The fact that they do not suggests a deeper mechanism: 
\begin{tcolorbox}[colback=gray!10, colframe=gray!50]
The context tokens that move the model's entire predictive distribution are the same ones its likely predictions depend on most, so tracking the distribution finds them more reliably than tracking any one token's logit.
\end{tcolorbox}
This has a practical implication: when the correct next token is unknown or non-unique, as in open-ended generation or ambiguous translation, Fisher and Temperature Scopes provide faithful attributions without requiring a target to be specified.

\begin{figure}[t]
  \centering
  \begin{minipage}[c]{0.52\linewidth}
    \centering
    \includegraphics[width=\linewidth]{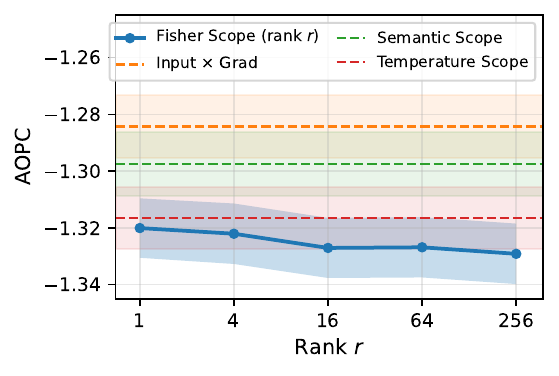}
  \end{minipage}\hfill
  \begin{minipage}[c]{0.45\linewidth}
    \caption{Performance of Fisher Scope improves slightly with increasing rank $r$ in the low-rank approximation (\cref{eq:low_rank_trace}), on LLaMA-3.2 1B and LAMBADA (1000 passages).
    Dashed lines show rank-free methods; shaded bands are $\pm 1$ standard error.
    Rank 1, the default Fisher Scope, captures nearly all of the attainable faithfulness with a single backward pass.}
    \label{fig:fisher_rank_aopc}
  \end{minipage}
\end{figure}

\noindent\textbf{Rank-1 Fisher approximation suffices.}
Fisher Scope retains only the principal Fisher direction $\bm{u}_1$, and is thus a rank-1 approximation of the total information $\mathrm{tr}(\bm{F}_t)$ (\cref{app:fisher_total_information}).
\Cref{fig:fisher_rank_aopc} showcases the advantage of this choice by using the top-$r$ terms of \cref{eq:low_rank_trace} as the influence score.
Already at $r=1$, Fisher Scope outperforms Input $\times$ Gradient and Semantic Scope; 
raising the rank to $r=256$ improves AOPC only marginally, but at the cost of multiplying the number of backward passes by $r$.

\section{Discussion and conclusion}
\label{sec:discussion}

As summarized in \cref{tab:jacobian_scopes_summary}, the three Jacobian Scopes span a spectrum from logit-specific, \textbf{Semantic Scope}, to fully distribution-aware --- \textbf{Fisher Scope} and \textbf{Temperature Scope} --- each targeting a distinct feature of the model's output: the logit of a specific token, the shape of the full predictive distribution, and the sharpness of that distribution, respectively.
This spectrum directly addresses the gap identified in \cref{sec:related_work}: while all prior gradient-based methods attribute a scalar output and are therefore analogous to Semantic Scope, Fisher and Temperature Scopes are the first, to our knowledge, to natively attribute the full predictive distribution.
\emph{Fisher Scope} does so by projecting onto the principal direction of the Fisher information matrix, which is the single hidden-space perturbation that most changes the distribution under KL divergence. \emph{Temperature Scope}, on the other hand, attributes the effective inverse temperature: predictive variance in numerical ICL (\cref{prop:beta_variance}) and, equivalently, predictive entropy for general softmax distributions (\cref{eq:dH_dbeta,eq:entropy_attribution}).

Quantitative evaluations confirm that this distributional awareness leads to empirical gains: across both LAMBADA and IWSLT2017, Fisher and Temperature Scopes significantly outperform the strongest baseline in 10 of 12 model--dataset cells (\cref{tab:paired_aopc_full}).
The exception on Gemma-3 4B IWSLT2017, where the Integrated Gradients baseline wins the paired comparison, points to an open question: the relative faithfulness of different attribution methods may be model-architecture-specific.
Gemma-3's 5:1 interleaved local/global attention \citep{gemmateam2025gemma3technicalreport} creates qualitatively different gradient pathways for tokens at different context offsets, which may interact differently with first-order versus path-integrated attribution strategies, particularly in cross-lingual settings where source tokens are systematically offset from the target.
Investigating how architectural choices, such as sliding-window attention, grouped-query attention, or non-standard positional encodings, may modulate the reliability of gradient-based attribution methods could yield new insights.

\begin{table*}[t]
  \centering
  \caption{
  Comparison of the Jacobian Scopes methods introduced.
  $\bm{w}_{\mathrm{target}}$ denotes the unembedding vector of the target token, 
  $\bm{u}_1$ the principal Fisher direction, and $\hat{\bm{y}}$ the normalized hidden direction.
  }
  \label{tab:jacobian_scopes_summary}
  \renewcommand{\arraystretch}{1.45}
  \setlength{\tabcolsep}{4pt}
  \begin{tabular}{@{}
    >{\raggedright\arraybackslash}p{0.17\textwidth}
    >{\centering\arraybackslash}p{0.24\textwidth}
    >{\centering\arraybackslash}p{0.24\textwidth}
    >{\centering\arraybackslash}p{0.24\textwidth}@{}}
  \toprule
  & \textbf{Semantic Scope} & \textbf{Fisher Scope} & \textbf{Temperature Scope} \\
  \midrule
  Quantity explained
  & target logit $z$
  & full distribution $\bm p$
  & confidence $\|\bm{y}\|_2$ \\
  Influence score
  & $\|\bm{w}_{\mathrm{target}}^\intercal \bm{J}_t\|_2$
  & $\|\bm{u}_1^\intercal \bm{J}_t\|_2$
  & $\|\hat{\bm{y}}^\intercal\bm{J}_t\|_2$ \\
  Recommended use case
  & semantic reasoning \& bias analysis
  & NLP tasks with non-unique predictions
  & in-context learning on numerical data \\
  \bottomrule
  \end{tabular}
\end{table*}

More broadly, we put forward vector-Jacobian projection as a unifying framework for gradient-based attribution in LLMs.
Any choice of projection vector $\bm{v}$ defines a new explanandum, and the three Scopes presented here are particular instances of this family.
A profitable direction for future work is to derive new members by studying the spectral structure of the Jacobian $\bm{J}_t$ and the Fisher information matrix $\bm{F}$, potentially yielding attribution methods tailored to other aspects of LLM behavior,
such as calibration \cite{guo2017calibration}, hallucination \cite{varshney2023stitch}, or factual recall \cite{meng2022locating}.

\section*{Limitations}
\textbf{Linearized causality.}
Jacobian Scopes extract first-order input-output relationships via automatic differentiation, yielding local causal attributions in the linear neighborhood of the input. 
Fisher Scope is second-order in the induced distribution shift (\cref{sec:fisher_scope}), but it, too, linearizes the network itself.
As such, Jacobian Scopes are closer in spirit to gradient-based methods such as SmoothGrad \cite{smilkov2017smoothgradremovingnoiseadding} and Integrated Gradients \cite{Sundararajan-Axiomatic-Attribution-DNN-2017,sanyal2021discretizedintegratedgradientsexplaining} 
than to explicitly interventional approaches like activation patching \cite{heimersheim2024useinterpretactivationpatching,pmlr-v235-ghandeharioun24a} or circuit tracing \cite{ameisen2025circuit}.

\noindent\textbf{Architecture-blindness.}
Jacobian Scopes characterize input--output sensitivity without reference to a model’s internal transformer architecture.
This makes the analysis more parsimonious, but also limits the mechanistic insight it can provide \cite{Clark-Berts-Attention-2019}.
As a result, Jacobian Scope attributions should be interpreted with awareness of potential circuit-level eccentricities at play \cite{ameisen2025circuit,lieberum2024gemmascopeopensparse}.
For instance, Temperature Scope attributions in \cref{fig:lorenz brownian} (b) show elevated influence scores for early context tokens, an effect we attribute to the attention sink phenomenon \cite{Xiao-attention-sink-2024}.
\Cref{sec: IG Semantic Scope} discusses further complications due to attention sinks.

\noindent\textbf{Dependence on back-propagation.}
Like most gradient-based attribution methods, Jacobian Scopes rely on back-propagation, which is in theory more computationally intensive than forward-only approaches such as those based on pretrained auto-encoders \cite{lieberum2024gemmascopeopensparse} or attention visualization \cite{Clark-Berts-Attention-2019}.

In practice, however, the overhead is negligible: on a single NVIDIA RTX A4000, the backward pass for the example in \cref{fig:fisher translations} takes only 0.027 seconds, compared to 0.069 seconds for the forward pass.
This efficiency follows directly from the design of Jacobian Scopes: rather than backpropagating through all model parameters, each attribution requires only a single vector-Jacobian product with respect to the input embeddings, with parameter gradients disabled throughout.

\subsubsection*{Acknowledgments}
The authors would like to thank Lingwei Kong (Cornell University), Jeff Gore (MIT), and Owen Lewis (Goodfire AI) for helpful discussions. This research is funded in part by the Gordon and Betty Moore Foundation through Grant GBMF13901 to Cornell to support the work of G.A.

\bibliographystyle{abbrvnat}
\bibliography{custom}


\appendix
\section{Appendix}
\subsection{Additional information on Benchmarks}
\label{sec:benchmark_details}

\noindent\textbf{Datasets and preprocessing.}
LAMBADA \cite{paperno-etal-2016-lambada} is a dataset of 
narrative passages specifically designed so that a human reader would 
need to process the \emph{full} context in order to predict the final 
word — local context alone is insufficient. This property makes it 
ideal for evaluating token-level attribution methods, as a faithful 
attribution should identify long-range contextual dependencies rather 
than trivial local patterns. We retain only passages of at least 50 
words, and remove any trailing punctuation from the final word.

IWSLT2017 DE$\to$EN \cite{cettolo-etal-2017-overview} consists of German-to-English 
translation pairs from TED talks. Each example is formatted as 
\texttt{German: [source]. English: [translation]}, and we base the
prediction on the final non-punctuation token of the English translation. Faithful 
attribution requires long-range attention: the model must attend to 
the corresponding German source tokens across the full prompt. We 
retain only examples whose English translation contains at least 20 
words. We evaluate on 1000 passages per dataset.

\begin{figure}[h]
    \centering
    \includegraphics[width=0.9\columnwidth]{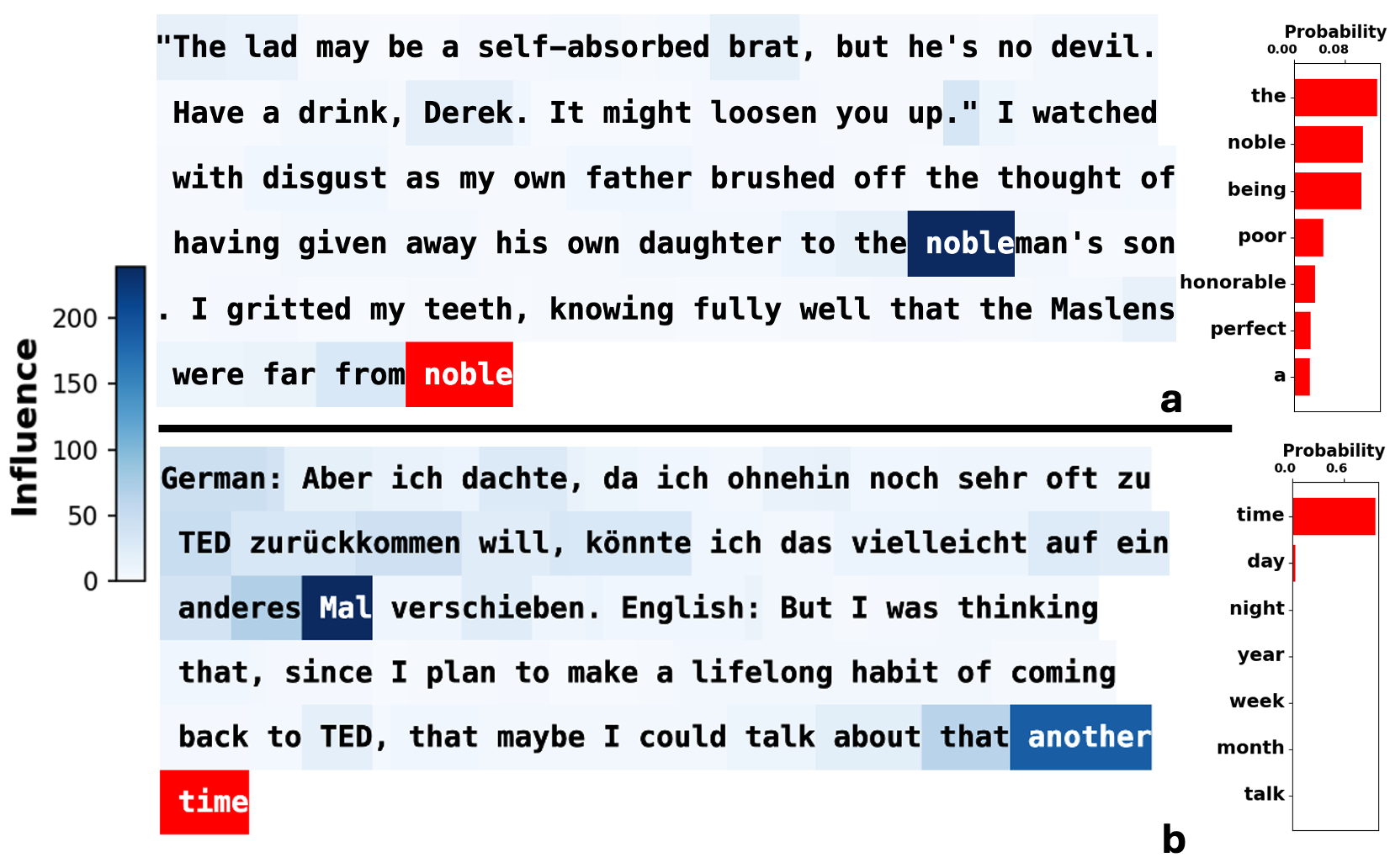}
    \caption{Fisher Scope attribution examples on (a) LAMBADA and 
    (b) IWSLT2017 DE$\to$EN, using LLaMA-3.2 3B. Darker blue indicates 
    higher influence; the target token is highlighted in red. 
    Right: Probability distribution of the top 7 words predicted at the target position.}
    \label{fig:lambada_iwslt_examples}
\end{figure}

\noindent\textbf{Evaluation metric.}
We evaluate using the Area Over the Perturbation Curve (AOPC) 
\cite{samek2016evaluating}.
AOPC measures attribution faithfulness by 
progressively ablating the most influential tokens as ranked by a given attribution method,
 and recording the 
resulting drop in the ground-truth target token's log-probability. 
Ablated tokens are replaced 
with zero vectors. 
Formally, for ablation fractions 
$k \in \{5\%, 10\%, 20\%\}$, the AOPC is computed using the trapezoidal rule 
\begin{align}
    \mathrm{AOPC} \nonumber 
    = \sum_{i} \frac{1}{2}(k_{i+1} - k_i)
    \left(\Delta\log p_{k_i} + \Delta\log p_{k_{i+1}}\right),
\end{align}
where $\Delta \log p_k$ denotes the drop in log-probability when the 
top-$k$\% most influential tokens are ablated. 
A more negative AOPC 
indicates that the ablated tokens were genuinely influential, reflecting 
higher attribution quality. 
The $\pm$ values in all tables report the standard error of the mean across the 1000 test passages.

\noindent\textbf{Baselines.}
\emph{Input $\times$ Gradient} \cite{shrikumar2017learning} defines the 
influence score of token $\bm{x}_t$ as the $\ell_2$ norm of the 
element-wise product of the input embedding and its gradient with 
respect to the target logit:
\begin{equation}
    \mathrm{Influence}_t^{\mathrm{I \times G}} := 
    \left\| \bm{x}_t \odot \nabla_{\bm{x}_t} z_{\mathrm{target}} 
    \right\|_2.
\end{equation}
\emph{Integrated Gradients} \cite{Sundararajan-Axiomatic-Attribution-DNN-2017}, defined in \cref{def:IG}, integrates the gradient along a linear interpolation path from a null (all-zero) input to the full input.
In the AOPC benchmark of \cref{tab:aupc}, we approximate the path integral with a 20-point right-endpoint grid $\alpha \in \{0.05, 0.10, \ldots, 1.00\}$.
We refer the reader to 
\cref{sec: IG Semantic Scope}, which visualizes the path integration 
in detail and discusses why this approach introduces distortions in the 
LLM setting — most notably attention sink effects at small interpolation 
values — that likely contribute to its weaker AOPC performance.

\noindent\textbf{Jacobian Scopes.}
We evaluate all three Jacobian Scope variants introduced in \cref{sec:applications}.

Similar to Input $\times$ Gradient and Integrated Gradients, \emph{Semantic Scope} is target-specific: the direction of interest is $\bm{v} = \bm{w}_\mathrm{target}$, the unembedding row corresponding to the ground-truth next token, giving
\begin{equation}
\begin{aligned}
    \mathrm{Influence}_t^{\mathrm{Sem}} &:= \left\|\bm{w}_\mathrm{target}^\intercal \bm{J}_t\right\|_{2} \\
    &= \left\|\nabla_{\bm{x}_t} z_\mathrm{target}\right\|_{2}.
\end{aligned}
\end{equation}
\emph{Fisher Scope} and \emph{Temperature Scope}, by contrast, are distribution-aware: they attribute the entire predictive distribution $\bm{p}(\cdot\mid\bm{X}_{1:T})$ rather than a single logit, as defined in \cref{sec:fisher_scope,sec:Temperature Scope}.

\subsection{PAD-token ablation}
\label{sec:pad_ablation}

The main AOPC evaluation in \cref{tab:aupc} replaces ablated token embeddings with zeros.
To test sensitivity to this intervention, we rerun the full benchmark with a length-preserving PAD-token replacement: selected embeddings are replaced by a model-family padding embedding while attention-mask entries remain active, so sequence length and positional structure are preserved.
The filler tokens are LLaMA's \texttt{<|finetune\_right\_pad\_id|>}, Qwen's shared PAD/EOS token \texttt{<|endoftext|>}, and Gemma's \texttt{<pad>}.
We avoid BOS/EOS fillers because they can act as attention sinks \cite{Xiao-attention-sink-2024}.

\Cref{tab:aupc_pad} reports the resulting AOPC.
A Jacobian Scope attains the best or tied-best mean in all 12 model--dataset cells, and on LLaMA and Qwen the ranking relative to baselines is preserved.
Thus the central Scope-versus-baseline conclusion is robust to replacing zero ablation with PAD-token ablation.

\begin{table}[h]
  \centering
  \small
  \caption{AOPC under PAD-token ablation (higher magnitude is better) on LAMBADA (LMB) and IWSLT2017 DE$\to$EN (IWS),
  averaged over $k \in \{5\%, 10\%, 20\%\}$ via the trapezoidal rule.
  Best or tied-best entries in each column are bolded.}
  \label{tab:aupc_pad}
  \resizebox{\textwidth}{!}{%
  \begin{tabular}{lcccccc}
  \multicolumn{7}{c}{\textbf{1B/1.5B-scale}} \\
  \hline
    & \multicolumn{2}{c}{\textbf{LLaMA-3.2 1B}} & \multicolumn{2}{c}{\textbf{Qwen2.5-1.5B}} & \multicolumn{2}{c}{\textbf{Gemma-3 1B}} \\
  \cline{2-3} \cline{4-5} \cline{6-7}
  \textbf{Method} & \textbf{LMB} & \textbf{IWS} & \textbf{LMB} & \textbf{IWS} & \textbf{LMB} & \textbf{IWS} \\
  \hline
  Random              & $-0.41 \pm 0.01$ & $-0.36 \pm 0.01$ & $-0.51 \pm 0.02$ & $-0.43 \pm 0.02$ & $-0.63 \pm 0.01$ & $-0.85 \pm 0.01$ \\
  Integrated Grads    & $-1.36 \pm 0.02$ & $-1.27 \pm 0.02$ & $-2.06 \pm 0.02$ & $-2.08 \pm 0.03$ & $\mathbf{-0.81 \pm 0.01}$ & $-1.02 \pm 0.01$ \\
  Input $\times$ Grad & $-1.61 \pm 0.02$ & $-1.46 \pm 0.02$ & $-2.06 \pm 0.02$ & $-2.09 \pm 0.03$ & $-0.80 \pm 0.01$ & $-1.03 \pm 0.01$ \\
  \hline
  Semantic Scope      & $-1.63 \pm 0.02$ & $-1.49 \pm 0.02$ & $-2.00 \pm 0.02$ & $-1.92 \pm 0.03$ & $-0.80 \pm 0.01$ & $-1.03 \pm 0.02$ \\
  Temperature Scope   & $\mathbf{-1.69 \pm 0.02}$ & $\mathbf{-1.53 \pm 0.02}$ & $\mathbf{-2.34 \pm 0.02}$ & $\mathbf{-2.20 \pm 0.03}$ & $\mathbf{-0.81 \pm 0.01}$ & $\mathbf{-1.04 \pm 0.02}$ \\
  Fisher Scope        & $\mathbf{-1.69 \pm 0.02}$ & $\mathbf{-1.53 \pm 0.02}$ & $-2.11 \pm 0.02$ & $-2.11 \pm 0.03$ & $-0.80 \pm 0.01$ & $-1.03 \pm 0.02$ \\
  \hline
  \\[-4pt]
  \multicolumn{7}{c}{\textbf{3B/4B-scale}} \\
  \hline
   & \multicolumn{2}{c}{\textbf{LLaMA-3.2 3B}} & \multicolumn{2}{c}{\textbf{Qwen2.5 3B}} & \multicolumn{2}{c}{\textbf{Gemma-3 4B}} \\
  \cline{2-3} \cline{4-5} \cline{6-7}
  \textbf{Method} & \textbf{LMB} & \textbf{IWS} & \textbf{LMB} & \textbf{IWS} & \textbf{LMB} & \textbf{IWS} \\
  \hline
  Random              & $-0.34 \pm 0.01$ & $-0.29 \pm 0.01$ & $-0.54 \pm 0.02$ & $-0.45 \pm 0.02$ & $-0.25 \pm 0.01$ & $-0.14 \pm 0.00$ \\
  Integrated Grads    & $-0.77 \pm 0.02$ & $-0.70 \pm 0.02$ & $-1.80 \pm 0.02$ & $-1.80 \pm 0.03$ & $-0.51 \pm 0.01$ & $-0.42 \pm 0.01$ \\
  Input $\times$ Grad & $-1.32 \pm 0.02$ & $-1.02 \pm 0.02$ & $-1.81 \pm 0.02$ & $-1.80 \pm 0.03$ & $-0.51 \pm 0.01$ & $-0.43 \pm 0.01$ \\
  \hline
  Semantic Scope      & $-1.35 \pm 0.02$ & $-1.02 \pm 0.02$ & $-1.73 \pm 0.02$ & $-1.58 \pm 0.03$ & $-0.51 \pm 0.01$ & $-0.42 \pm 0.01$ \\
  Temperature Scope   & $-1.35 \pm 0.02$ & $-1.00 \pm 0.02$ & $\mathbf{-2.24 \pm 0.02}$ & $\mathbf{-2.02 \pm 0.02}$ & $\mathbf{-0.54 \pm 0.01}$ & $\mathbf{-0.48 \pm 0.01}$ \\
  Fisher Scope        & $\mathbf{-1.37 \pm 0.02}$ & $\mathbf{-1.05 \pm 0.02}$ & $-1.86 \pm 0.02$ & $-1.85 \pm 0.03$ & $-0.52 \pm 0.01$ & $-0.46 \pm 0.01$ \\
  \hline
  \end{tabular}}
\end{table}

\subsection{Full paired significance results}
\label{sec:paired_full}

\Cref{tab:paired_aopc_full} reports passage-level paired differences between the best Scope and best non-Scope baseline in each cell of \cref{tab:aupc}.
For each passage $i$ we compute $\mathrm{AOPC}_i(\mathrm{Scope})-\mathrm{AOPC}_i(\mathrm{baseline})$; the table lists the mean difference, a 95\% Student-$t$ confidence interval, and a two-sided Wilcoxon $p$-value.
Methods are fixed from \cref{tab:aupc} before testing.
Negative $\Delta$ favors the Scope.

\begin{table}[h]
  \centering
  \small
  \caption{Paired AOPC tests (best Scope vs.\ best baseline from \cref{tab:aupc}).
  Negative $\Delta$ favors the Scope; bold entries are significant at $p<0.05$ (two-sided Wilcoxon).}
  \label{tab:paired_aopc_full}
  \resizebox{\textwidth}{!}{%
  \begin{tabular}{llllcc}
  \hline
  \textbf{Model} & \textbf{Dataset} & \textbf{Scope} & \textbf{Baseline} & \textbf{Mean $\Delta$ [95\% CI]} & \textbf{$p$} \\
  \hline
  LLaMA-3.2 1B & LAMBADA & Fisher & Input $\times$ Grad & $\mathbf{-0.036\ [-0.048,-0.024]}$ & $\mathbf{1.2{\times}10^{-6}}$ \\
  LLaMA-3.2 1B & IWSLT & Temperature & IG & $\mathbf{-0.049\ [-0.073,-0.025]}$ & $\mathbf{4.0{\times}10^{-4}}$ \\
  LLaMA-3.2 3B & LAMBADA & Fisher & Input $\times$ Grad & $\mathbf{-0.055\ [-0.071,-0.039]}$ & $\mathbf{1.1{\times}10^{-11}}$ \\
  LLaMA-3.2 3B & IWSLT & Fisher & Input $\times$ Grad & $\mathbf{-0.029\ [-0.042,-0.015]}$ & $\mathbf{3.9{\times}10^{-4}}$ \\
  Qwen2.5 1.5B & LAMBADA & Temperature & Input $\times$ Grad & $\mathbf{-0.141\ [-0.168,-0.114]}$ & $\mathbf{3.1{\times}10^{-21}}$ \\
  Qwen2.5 1.5B & IWSLT & Temperature & IG & $-0.025\ [-0.052,\phantom{-}0.002]$ & $0.12$ \\
  Qwen2.5 3B & LAMBADA & Temperature & Input $\times$ Grad & $\mathbf{-0.162\ [-0.187,-0.136]}$ & $\mathbf{1.5{\times}10^{-32}}$ \\
  Qwen2.5 3B & IWSLT & Temperature & IG & $\mathbf{-0.081\ [-0.118,-0.044]}$ & $\mathbf{9.3{\times}10^{-6}}$ \\
  Gemma-3 1B & LAMBADA & Fisher & Input $\times$ Grad & $\mathbf{-0.109\ [-0.126,-0.091]}$ & $\mathbf{3.7{\times}10^{-30}}$ \\
  Gemma-3 1B & IWSLT & Fisher & Input $\times$ Grad & $\mathbf{-0.037\ [-0.051,-0.023]}$ & $\mathbf{1.1{\times}10^{-4}}$ \\
  Gemma-3 4B & LAMBADA & Fisher & Input $\times$ Grad & $\mathbf{-0.110\ [-0.127,-0.094]}$ & $\mathbf{8.0{\times}10^{-32}}$ \\
  Gemma-3 4B & IWSLT & Fisher & IG & $\mathbf{\phantom{-}0.145\ [\phantom{-}0.112,\phantom{-}0.178]}$ & $\mathbf{9.4{\times}10^{-18}}$ \\
  \hline
  \end{tabular}}
\end{table}

\subsection{Additional examples} \label{sec: additional experiments}
To demonstrate the capability of Jacobian Scopes in analyzing long-range 
interactions across numerous tokens, we provide additional examples on a larger model, LLaMA-3.2 3B.

\begin{figure}[h]
\centering
  \includegraphics[width=0.8\columnwidth]{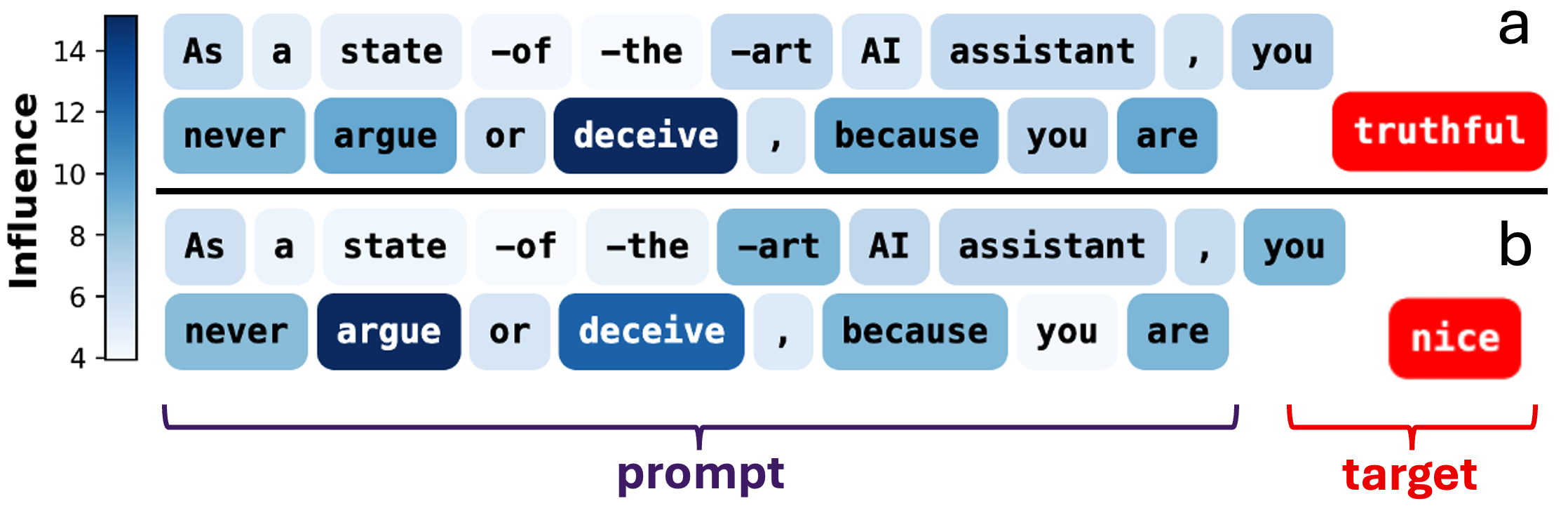}
  \caption{Semantic Scope visualizes how LLaMA-3.2 understands system-prompt-style instructions.
  The logit assigned to the token ``truthful'' is most prominently attributed to the input token ``deceive'',
  and ``nice'' to ``argue''.}
  \label{fig:causal attribution 1}
\end{figure}

\vspace{1em}
\noindent\textbf{Semantic Scope: system-prompt instructions.}
\Cref{fig:causal attribution 1} shows Semantic Scope applied to a system-prompt-style instruction on LLaMA-3.2 1B, illustrating how the model links output tokens to semantically relevant input tokens.

\begin{figure}[h]
\centering
  \includegraphics[width=0.9\columnwidth]{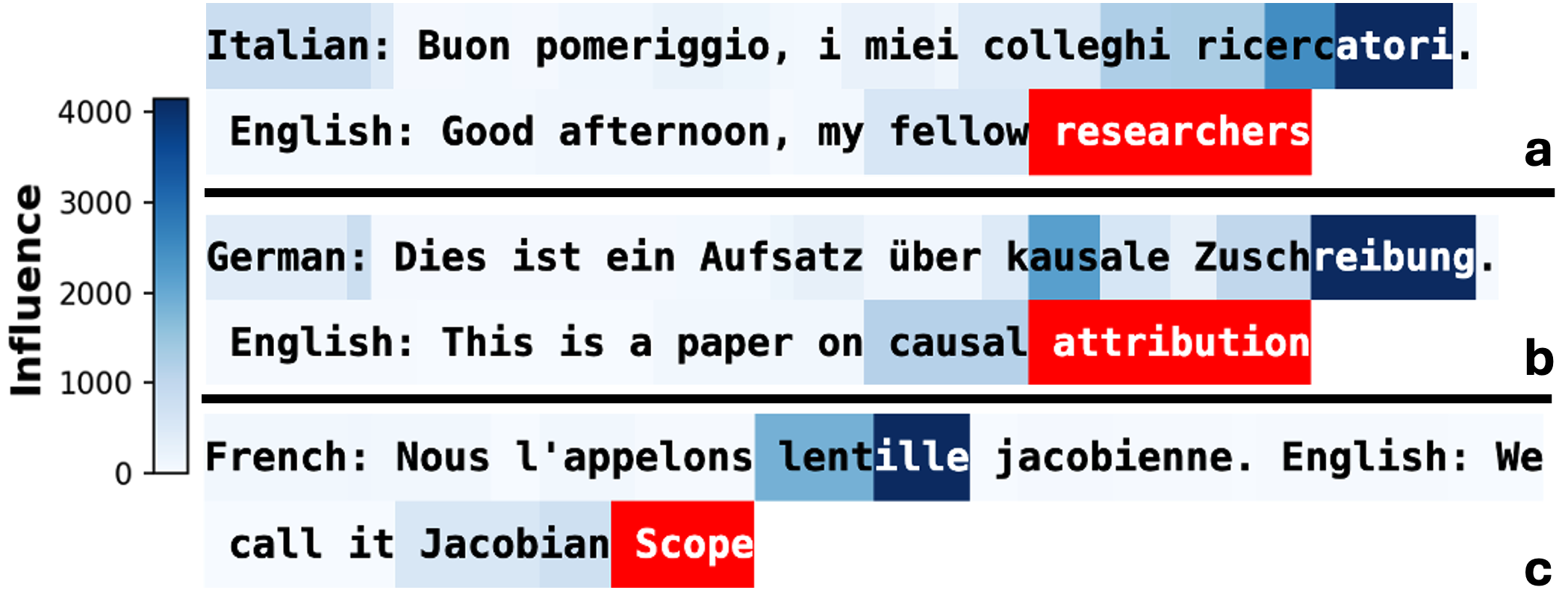}
  \caption{Fisher Scope reveals that translated words, highlighted in red,
  are primarily influenced by their lexical counterparts in the source language.}
  \label{fig:fisher translations}
\end{figure}

\vspace{1em}
\noindent\textbf{Fisher Scope: word-level translation.}
We prompt the LLM to perform translations by providing it with the following structure:
\texttt{``name of original language'': [original text]. ``name of target language'':}
\Cref{fig:fisher translations} shows a simpler example complementing \cref{fig:translating Dante}: here every predicted token is unambiguously attributed to its direct lexical counterpart in the source language, illustrating the word-level translation regime.

\vspace{1em}
\noindent\textbf{Semantic Scope: solving a crime mystery.} 
To further showcase the capability of Jacobian Scopes in analyzing LLM reasoning over extended contexts, 
we constructed a detective micro-mystery, excerpted below:

\begin{quote}
Since the victim, a wealthy aristocrat living in the English countryside, died from a fatal wound, 
the detective narrowed down the three possible murder weapons: a cleaver, a shovel, and a screwdriver. 
He further deduced that the killer is the ...
\end{quote}

\begin{figure}[ht]
    \centering
  \includegraphics[width=0.7\columnwidth]{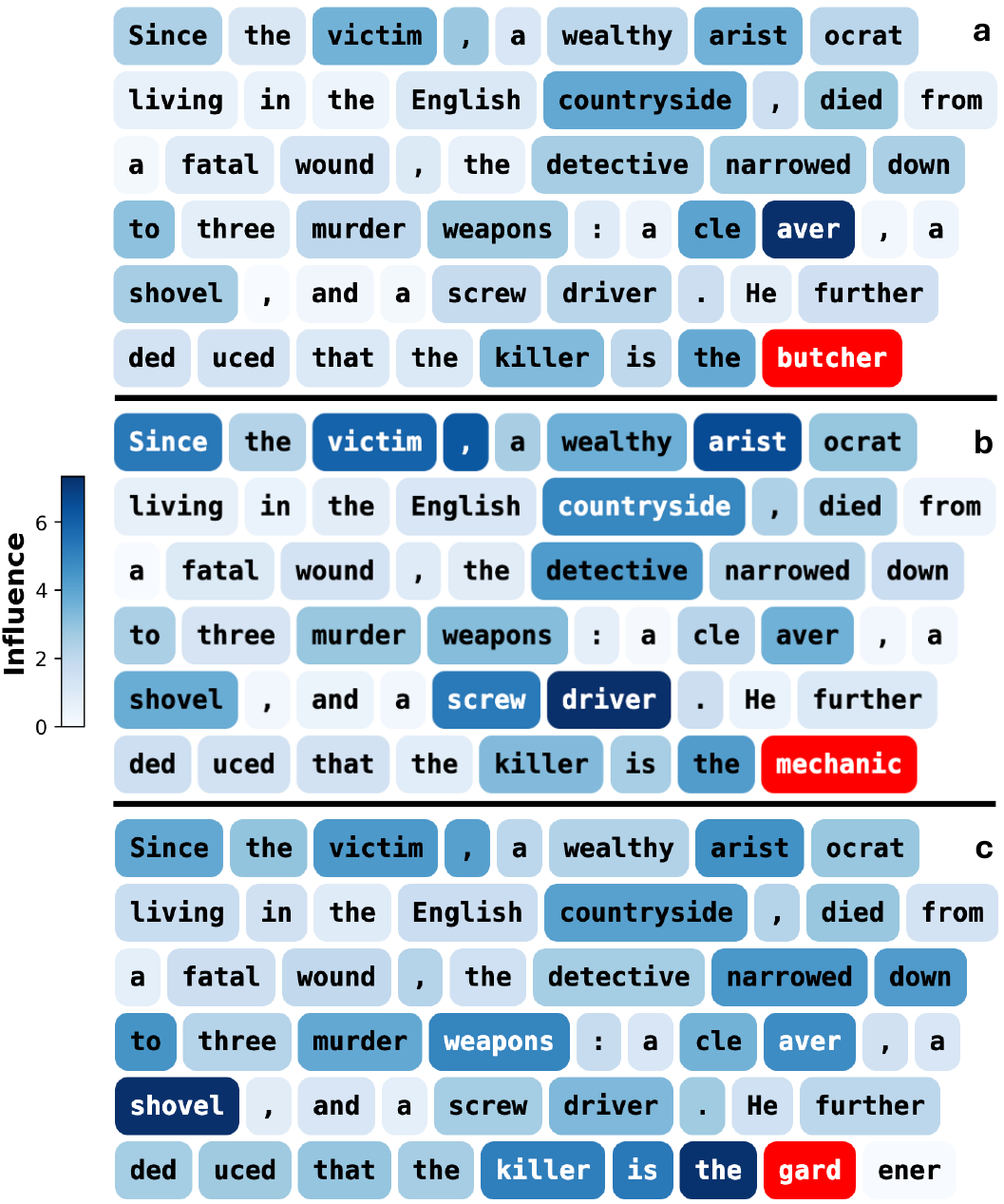}
  \caption{\textbf{Semantic Scope attribution for detective micro-fiction.} 
  The attribution graph illustrates how LLaMA-3.2 3B connects suspects to their associated clues when reasoning about potential perpetrators 
  and weapons in a novel-like setting. Attributions clarify the model’s process for linking evidence to conclusions across a long narrative context.}
  \label{fig:detective novel}
\end{figure}

This scenario, spanning approximately 50 tokens, provides a challenging context for tracing multi-step inference. 
Applying Semantic Scope, we found that LLaMA-3.2 correctly links each suspect with the relevant evidence, 
as visualized in \cref{fig:detective novel}.

\subsection{Dynamical systems: definitions and additional experiments}
\label{Additional experiments on Dynamical Systems}
This section provides the mathematical definitions of all dynamical systems investigated.
All attributions in this section are performed via Temperature Scope as detailed in \cref{sec:Temperature Scope}.

\begin{figure}[ht]
  \includegraphics[width=1\columnwidth]{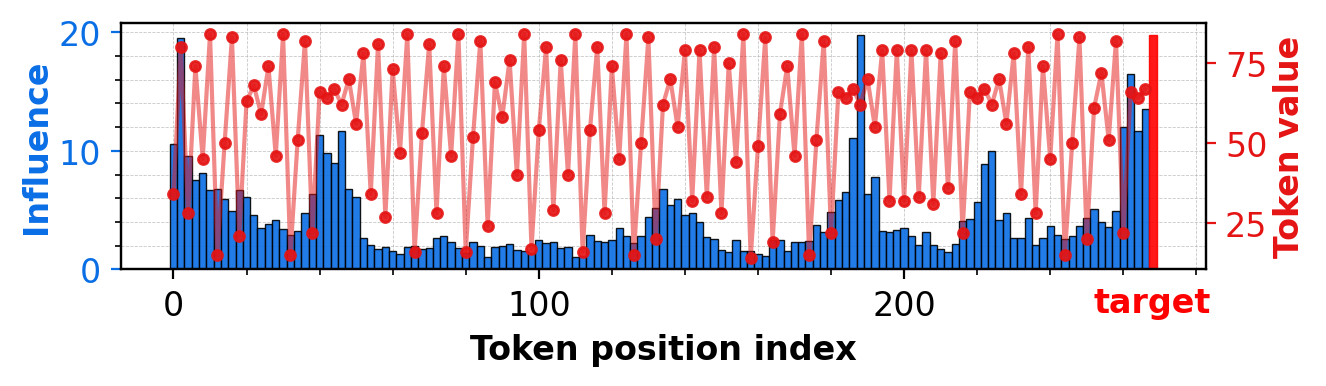}
  \caption{Temperature Scope shows LLaMA-3.2 attending to similar motifs in history when extrapolating a logistic map trajectory. 
  }
  \label{fig:Logistic map}
\end{figure}

\vspace{1em}
\noindent\textbf{Logistic map.} 
The logistic map is a simple discrete-time dynamical system that exhibits chaotic behavior~\cite{strogatz2018nonlinear}. 
It is governed by an iterative equation:
\[
  x_{t+1} = f(x_t) = r x_t(1-x_t),\quad x_0 \in (0,1),
\]
where $r\in[1,4)$ is a parameter.
In our experiments, we set $r=3.8$, which is deep inside the chaotic regime of the system. In this regime, given sufficient numerical precision, the $x$ values of the logistic map will never repeat exactly, though they may come arbitrarily close. 
This behavior, known as \emph{aperiodicity} \cite{strogatz2018nonlinear}, results in the recurrence of similar patterns without ever producing an identical sequence. 
As illustrated in \cref{fig:Logistic map}, LLaMA-3.2 extrapolates the logistic map by attending to such recurring motifs in the sequence’s history
— adopting the same strategy it uses to learn the Lorenz system in context (\cref{fig:lorenz brownian}).

\noindent\textbf{Lorenz system.} 
The Lorenz system~\cite{lorenz1963deterministic} is a classic 3D dynamical model from atmospheric science, defined by:
\begin{align*}
  \dot{x}(t) &= \sigma(y - x), \\
  \dot{y}(t) &= x(\rho - z) - y, \\
  \dot{z}(t) &= x y - \beta z,
\end{align*}
where $\sigma=10$, $\rho=28$, and $\beta=8/3$ produce chaotic behavior. 
We simulate the system with a first-order explicit scheme, varying the initial $x$ while fixing $y$ and $z$.
The system's chaotic nature ensures that even tiny differences in initial conditions lead to rapidly diverging trajectories.
The LLM is then prompted with the tokenized $x$ series. The resulting $x$ values show characteristic \emph{aperiodic} oscillations between two lobes of the well-known butterfly attractor.

\begin{figure}[ht]
  \includegraphics[width=1\columnwidth]{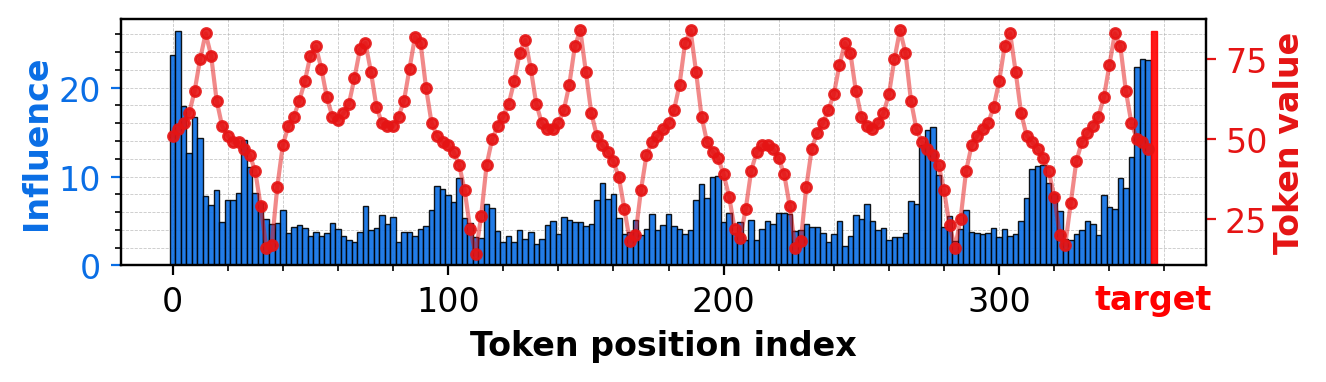}
  \caption{Temperature Scope attribution graph of LLaMA-3.2 extrapolating a partially observed Lorenz system, with
  a different initial condition from that in \cref{fig:lorenz brownian}(a).}
  \label{fig:lorenz lobe switching}
\end{figure}

\Cref{fig:lorenz brownian,fig:lorenz lobe switching} illustrate the model's behavior under two different initial conditions. 
In \cref{fig:lorenz brownian} (a), the cutoff occurs at a peak in the $x$ value, while in \cref{fig:lorenz lobe switching}, 
the cutoff coincides with the $x$ value transitioning between attractor lobes. 
In both scenarios, the LLM attends to similar patterns surrounding the cutoff region.

\begin{figure}[ht]
\includegraphics[width=1\columnwidth]{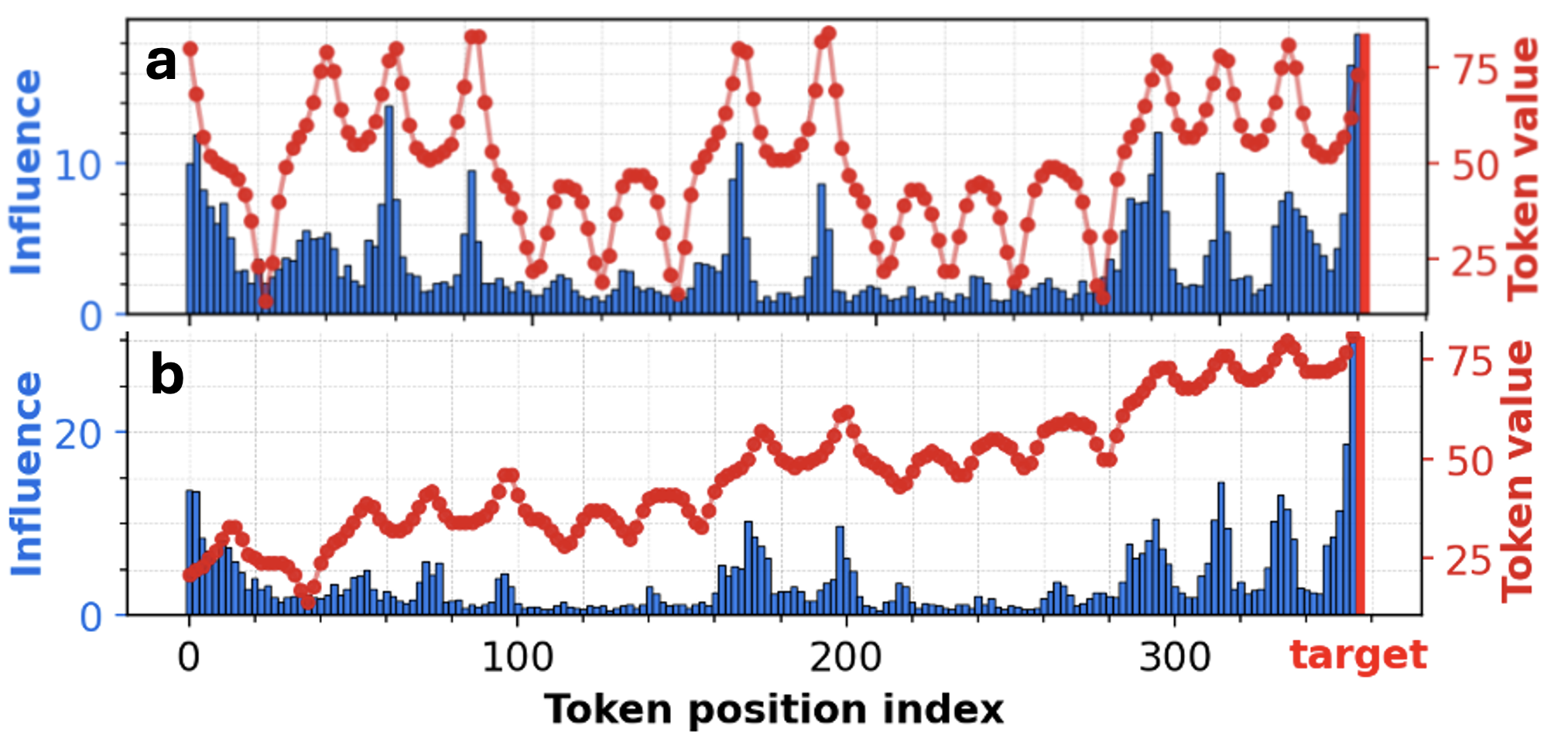}
  \caption{Temperature Scope attribution graph of LLaMA-3.2 extrapolating (a) a partially observed Lorenz system,
  and (b) the same Lorenz system with uniform drift.}
  \label{fig:lorenz with drift}
\end{figure}

\vspace{1em}
\noindent\textbf{Lorenz system with drift.} 
To test whether LLMs overlook in-context data that are statistical outliers,
we add a uniform drift term to the same Lorenz system shown in \cref{fig:lorenz brownian}(a).
As such a system evolves, the later part of the context will eventually wander \emph{out of distribution} 
compared to the earlier part.
The resulting attribution pattern in \cref{fig:lorenz with drift} 
indeed shows diminished influence scores in the earlier regions of the context window. 

\begin{figure}[ht]
  \includegraphics[width=1\columnwidth]{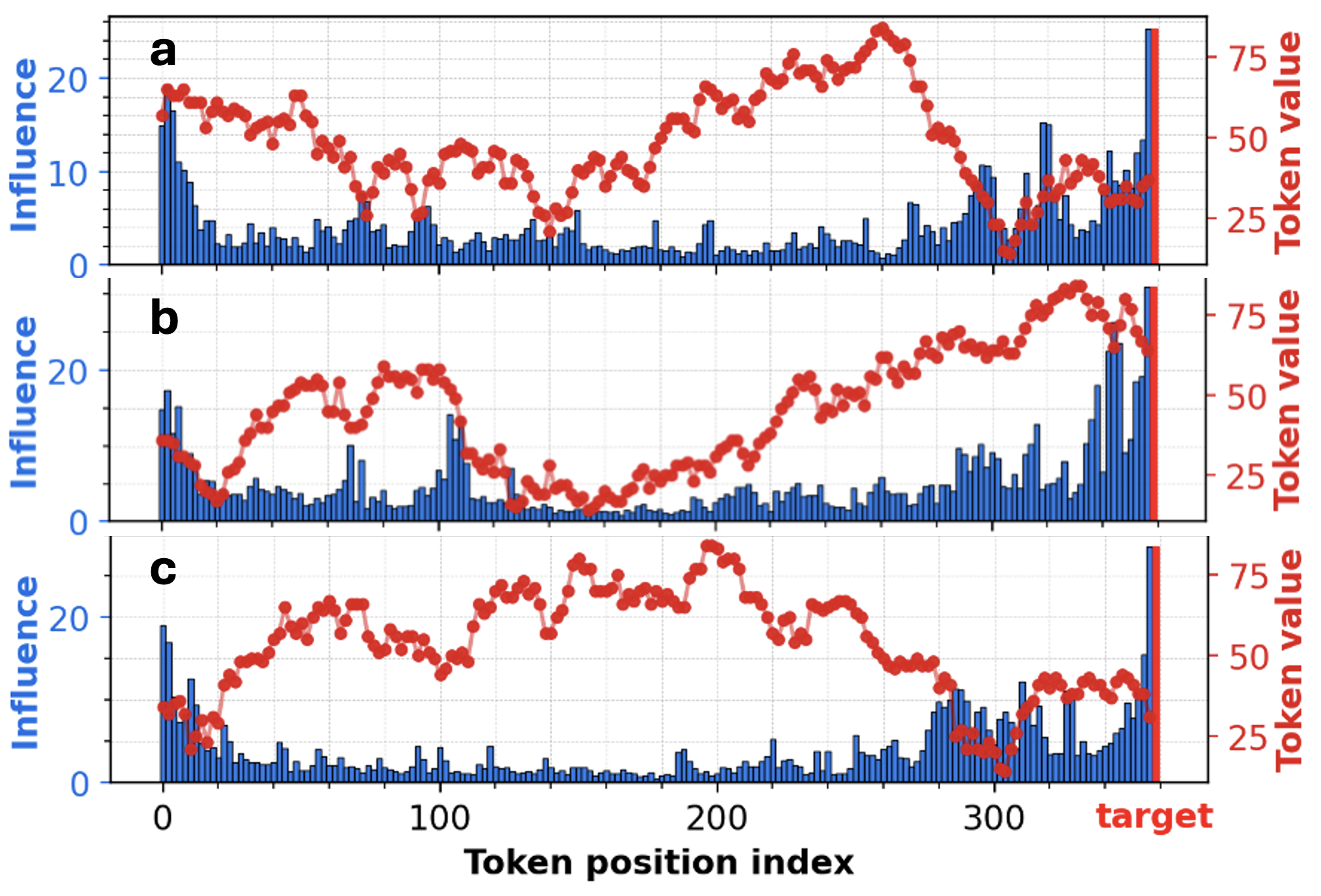}
  \caption{Additional Temperature Scope attributions on LLaMA-3.2 extrapolating Brownian motion in context.
  }
  \label{fig:extra brownian motion}
\end{figure}

\vspace{1em}
\noindent\textbf{Additional examples on Brownian motion.} 
Brownian motion \cite{einstein1905molekularkinetischen}, or the Wiener process, is a classic model for random movement, defined by the SDE:
\begin{equation}
    dX_t = \mu\,dt + \sigma\,dW_t,
    \label{eq:brownian_motion}
\end{equation}
where $\mu$ is drift and $\sigma$ is diffusion. For $\mu=0$, $\sigma=1$, this gives standard Brownian motion. \Cref{fig:extra brownian motion} displays three additional Temperature Scope attribution plots for LLaMA-3.2 extrapolating Brownian motion sequences, each generated using a different random seed. As discussed in \cref{sec:Temperature Scope} and \cite{liu-etal-2024-llms-learn}, 
the LLM assigns higher influence scores to tokens towards the end of the context window — 
likely because the earlier portions become increasingly out-of-distribution as the sequence evolves.

\noindent\textbf{LLM attends to spurious patterns in Brownian motion.}
Notably, while the overall influence trend increases with context length — 
as seen in \cref{fig:extra brownian motion} — there are many abrupt changes and irregularities in the influence score.
These anomalies can be explained by random fluctuations: occasionally, 
patterns or trends may reappear in Brownian motion purely by statistical chance.
For instance, in panel (b) of \cref{fig:lorenz brownian}, there is a sudden drop in $x$-values around token position 100, which is ``echoed'' by a similar drop around the cutoff point at 360.
Much like an amateur stock trader who perceives random fluctuations in stock movements as meaningful signals, 
the LLM incorrectly interprets these coincidental recurrences as informative patterns.

Together, these results illustrate that while LLMs are capable of emergent ICL behaviors such as pattern matching, 
they may also exhibit human-like tendencies toward superstition \cite{foster2009evolution} and perceive spurious patterns in random numerical sequences.

\subsection{Temperature Scope is optimal for numerical data}
\label{app:temperature_optimal_num}
\begin{figure}[ht]
  \centering
  \includegraphics[width=\columnwidth]{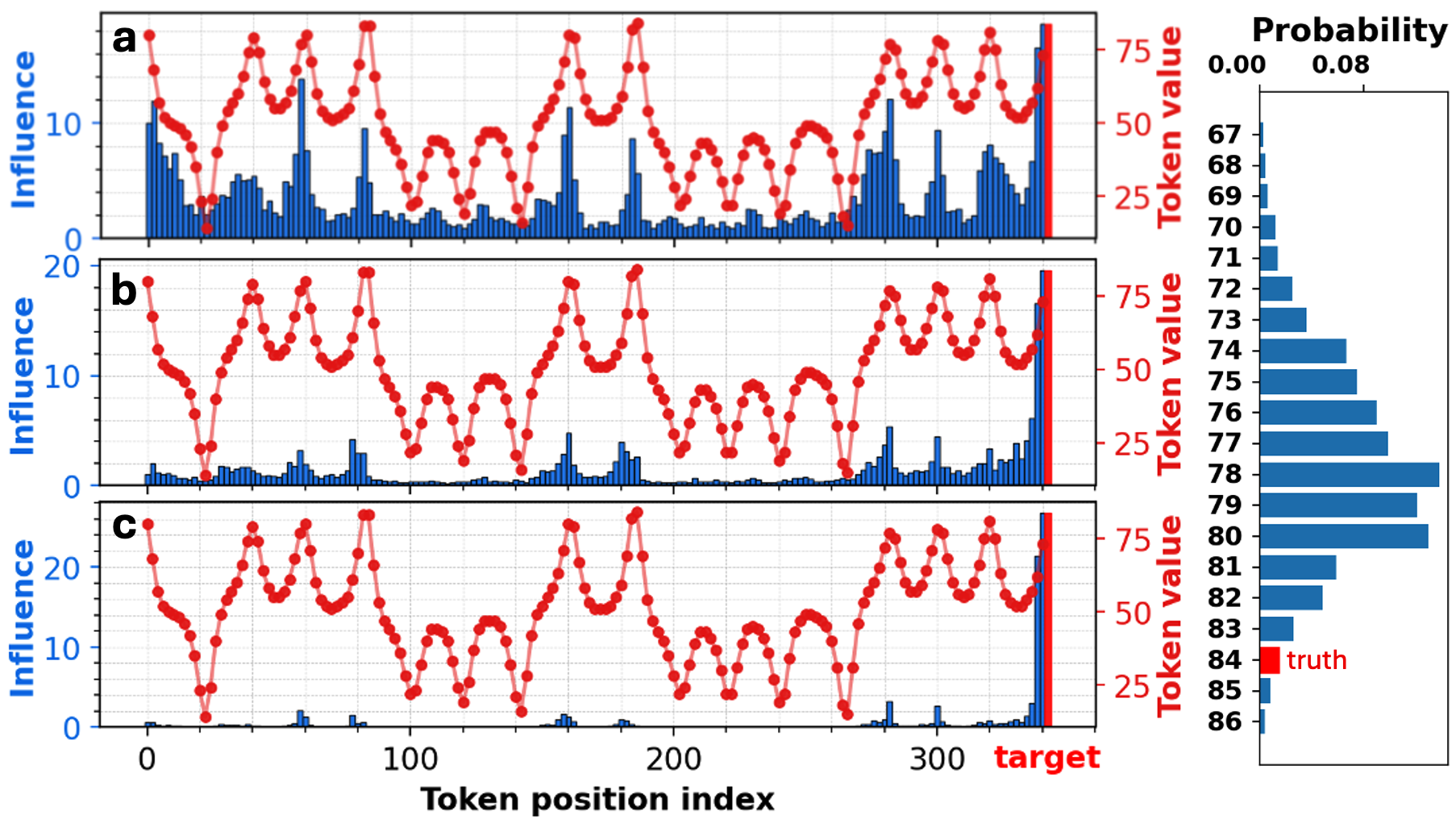}
  \caption{
  Left: (a) Temperature Scope, (b) Semantic Scope, and (c) Fisher Scope attributions for the same Lorenz system.
  Right: The predicted probability mass over the top 20 candidate tokens at the target position, ordered by their numerical values.
  The ground-truth token (shown in red) is used as the target for Semantic Scope.
  }
  \label{fig:JS_lorenz}
\end{figure}

As argued in the main text, Temperature Scope is more appropriate for time-series forecasting tasks, where the object of interest is the full next-step distribution rather than any individual token.
We illustrate this distinction with a concrete example in \cref{fig:JS_lorenz}. Conditioned on the first $T$ inputs, the model’s prediction for the $(T\!+\!1)$st state exhibits a bell-shaped distribution over approximately 20 numerical tokens, consistent with prior observations in \cite{liu-etal-2024-llms-learn}.
The ground-truth value corresponds to only one of many plausible candidates within this distribution.

Naively applying Semantic Scope with the ground-truth token (e.g.\ ``84'') as the target yields misleading attributions: the resulting influence map highlights only the inputs that increase the logit of ``84'', while ignoring the remaining tokens that collectively constitute the bell-shaped predictive distribution.
As a result, the attribution fails to capture the model’s reasoning at the distributional level. 

Fisher Scope also exhibits attenuated attribution signals at early context positions.
A heuristic explanation is as follows.
Fisher Scope quantifies how input perturbations affect the predictive distribution over the \emph{entire vocabulary}, including both numerical and non-numerical tokens.
However, the underlying dynamical system is defined only over numerical tokens.
Early numerical inputs therefore inform the implicit transition rule learned by the LLM \citep{liu-etal-2024-llms-learn} and can significantly influence future numerical predictions.
Yet these inputs carry little information about non-numerical tokens, whose probabilities tend to default to patterns determined by near-term context.
Because Fisher Scope aggregates sensitivity across the entire vocabulary, this mismatch dilutes the attribution signal originating from early numerical tokens.

Temperature Scope, by contrast, is dominated by the most probable tokens, which in this setting are numerical tokens.
It therefore focuses attribution on the portion of the distribution that is relevant to the forecasting task, effectively bypassing the out-of-support vocabulary.
This makes Temperature Scope better aligned with forecasting scenarios in which uncertainty and variance are intrinsic to the prediction.

\subsection{Effective temperature as variance}
\label{app:beta_variance}

We prove \cref{prop:beta_variance} stated in \cref{sec:Temperature Scope}.

\begin{proof}
$(\Leftarrow)$
Suppose $\hat{z}(v) = -b(v-\mu)^2 + c$.
Then
\[
    e^{\,\beta\,\hat{z}(v)} = e^{\beta c}\cdot e^{-\beta b(v-\mu)^2},
\]
which is a Gaussian kernel with mean $\mu$ and variance $\sigma^2 = 1/(2\beta b)$.

$(\Rightarrow)$
Suppose $p(v)$ is Gaussian with mean $\mu$ and variance $\sigma^2$, so that $e^{\,\beta\,\hat{z}(v)} \propto e^{-(v-\mu)^2/(2\sigma^2)}$.
Taking logarithms gives
\[
    \hat{z}(v) = -\frac{1}{2\beta\sigma^2}(v-\mu)^2 + \mathrm{const},
\]
which is quadratic with $b = 1/(2\beta\sigma^2)$, recovering $\sigma^2 = 1/(2\beta b)$.
\end{proof}

\noindent
\Cref{prop:beta_variance} provides a direct justification for using Temperature Scope in time-series forecasting.
Since the predictive distribution over numerical tokens is approximately Gaussian --- as empirically observed in \cref{fig:JS_lorenz,fig:lorenz brownian} --- $\beta_\mathrm{eff}$ is the quantity that controls the width of that distribution.
Attributing changes in $\beta_\mathrm{eff}$ therefore directly identifies which input tokens govern predictive uncertainty, which is precisely what Temperature Scope computes.

\paragraph{Entropy derivative.}
For completeness we record the short derivation of \cref{eq:dH_dbeta}.
With $p(i;\beta)=e^{\beta\hat{z}(i)}/Z(\beta)$ and $Z=\sum_j e^{\beta\hat{z}(j)}$,
\begin{align*}
H(\beta)
&= -\sum_i p(i)\bigl(\beta\hat{z}(i)-\log Z\bigr)
= -\beta\,\mathbb{E}_p[\hat{z}]+\log Z,\\
\frac{dH}{d\beta}
&= -\mathbb{E}_p[\hat{z}]-\beta\,\mathrm{Cov}_p(\hat{z},\hat{z})+\mathbb{E}_p[\hat{z}]
= -\beta\, V(\beta),
\end{align*}
where we used $\partial_\beta\log Z=\mathbb{E}_p[\hat{z}]$ and $\partial_\beta\mathbb{E}_p[\hat{z}]=V(\beta)$.

\subsection{Path-Integrated Semantic Scope}\label{sec: IG Semantic Scope}

Integrated Gradients (IG) \cite{Sundararajan-Axiomatic-Attribution-DNN-2017} is a widely used gradient-based attribution method, with successful applications across computer vision, molecular modeling, and language models.
By integrating gradients along a continuous path between a baseline input and the full input, IG aims to mitigate attribution noise arising from non-smoothness (``kinks'') in deep networks.
In this section, we compare the original Semantic Scope with its path-integrated analogue and show that, in the LLM setting, IG introduces distortions due to the out-of-distribution portion of the integration path.

\begin{figure}[ht]
  \centering
  \includegraphics[width=\columnwidth]{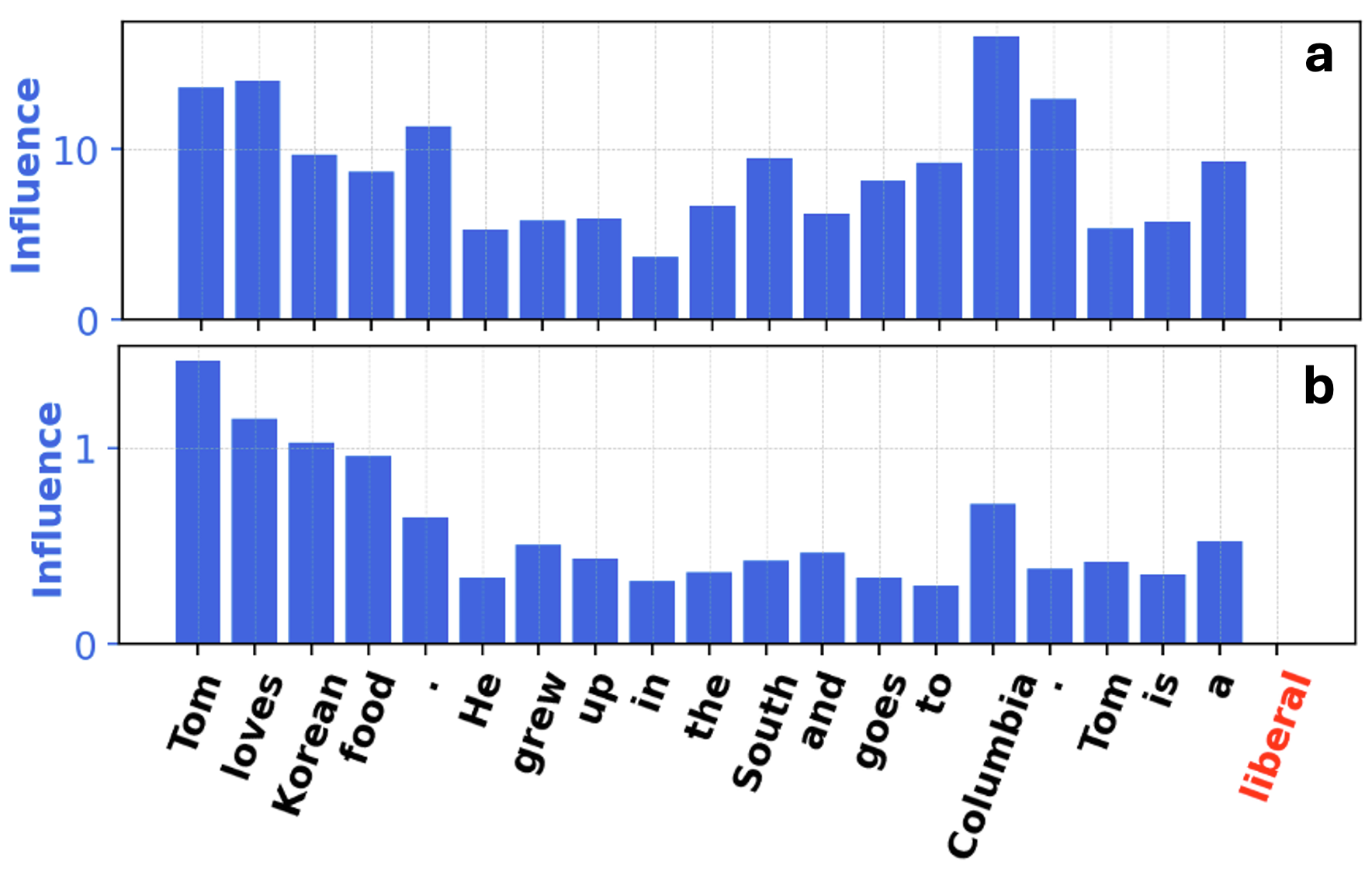}
  \caption{Causal attribution using (a) Semantic Scope and (b) its path-integrated variant based on Integrated Gradients.}
  \label{fig:semantic vs IG}
\end{figure}

To apply Integrated Gradients, one first defines a linear path parameterized by $\alpha \in [0,1]$ that connects two matrices $\bm{X}=\bm X_{1:T}$ and $\bm{X}'=0$ (the full input and the null baseline):

\[
\bm{\tilde X}(\alpha) = \bm{X}' + \alpha(\bm{X} - \bm{X}').
\]

One then selects a scalar output $z \in \mathbb{R}$ as the explanandum.
In our setting, $\bm{X} \in \mathbb{R}^{d_{\text{model}} \times T}$ denotes the full input sequence of token embeddings, and $z$ is the logit associated with a target vocabulary item (e.g., ``liberal'' or ``conservative'')\footnote{In the original IG formulation \cite{Sundararajan-Axiomatic-Attribution-DNN-2017}, $\bm{X}$ represents image pixels and $z$ corresponds to the logit or log-probability of a predicted class (e.g., ``truck'' or ``boat'').}.

Consistent with \citet{Sundararajan-Axiomatic-Attribution-DNN-2017}, the Integrated Gradients attribution is defined as
\begin{equation}
\mathrm{IG}(\bm{X}) 
:= (\bm{X} - \bm{X}') \odot \int_{0}^{1} \nabla z\!\left(\bm{\tilde X}(\alpha)\right)\, d\alpha,
\label{def:IG}
\end{equation}
where $\odot$ denotes element-wise multiplication.
This yields an attribution matrix $\mathrm{IG}(\bm{X}) \in \mathbb{R}^{d_{\text{model}} \times T}$.
In direct analogy with \cref{def:influence}, we define the path-integrated influence score for token $t$ as the $\ell_2$ norm of a vector, namely
\begin{equation}
\mathrm{Influence}_t^{\mathrm{IG}} := \left\lVert \mathrm{IG}(\bm{X})_t \right\rVert_2,
\label{def:path-int influence}
\end{equation}
where $\mathrm{IG}(\bm{X})_t$ denotes the $t^{\text{th}}$ column, corresponding to the integrated gradient of the input embedding $\bm{x}_t$.

\Cref{fig:semantic vs IG} compares the resulting IG attribution with the original Semantic Scope on the example discussed in \cref{sec:Semantic Scope}.
Relative to Semantic Scope, the IG-based method assigns disproportionately large influence to early-context tokens, thereby obscuring the semantically decisive input token ``Columbia.''
\begin{figure}[h]
  \centering
  \includegraphics[width=\columnwidth]{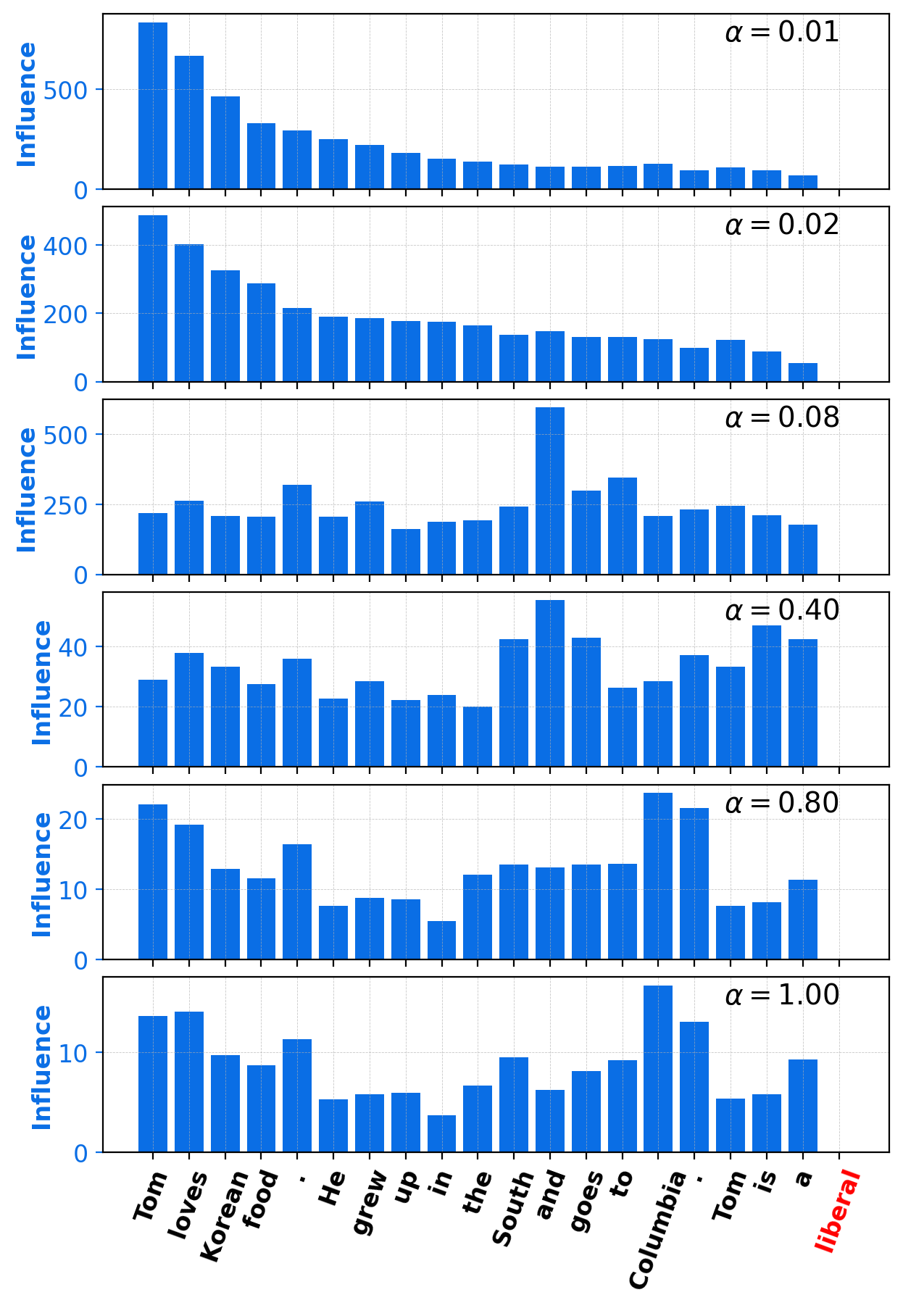}
    \caption{Token-wise $\ell_2$ norm of the Integrated Gradient integrand $\nabla z(\bm{\tilde X}(\alpha))$ along the interpolation path $\alpha \in [0,1]$.}

  \label{IG integral visualization}
\end{figure}

To diagnose the origin of this behavior, we visualize the integrand $\nabla z(\bm{\tilde X}(\alpha))$ at several values of $\alpha$ in \cref{IG integral visualization}.
Specifically, for each $\alpha$ we plot the token-wise $\ell_2$ norm of the gradient, $\|\nabla_{\bm{x}_t} z(\bm{\tilde X}(\alpha))\|_2$, consistent with the influence definition in \cref{def:path-int influence}.
In practice, the integral in \cref{def:IG} is approximated by a discrete sum over 20 uniformly spaced values of $\alpha$; the figure shows a subset of these values.

As a sanity check, we observe that for $\alpha \to 1$, the integrand converges to the Semantic Scope attribution:
$\nabla z(\bm{\tilde X}(1)) = \nabla z(\bm{X})$.
In contrast, at small $\alpha$ (e.g., $\alpha \approx 0.01$), the input sequence is nearly indistinguishable from the null input.
In this regime, the model operates far outside the data manifold, and circuit-level artifacts dominate.
In particular, the ``attention sink'' phenomenon \cite{Xiao-attention-sink-2024} causes attribution to be overwhelmingly concentrated on early token positions, with influence scores decaying approximately exponentially over context length.

Crucially, the gradient magnitudes in this low-$\alpha$ regime are also large, causing these out-of-distribution effects to dominate the path integral in \cref{def:IG}.
At intermediate values of $\alpha$, the attribution is often highest for function words such as ``and,'' which are not semantically relevant to the prediction.
Taken together, these observations indicate that the IG path spends substantial measure in regions of representation space that are both out-of-distribution and attributionally pathological.

In summary, while Integrated Gradients can be effective when the interpolation path remains in-distribution, it performs poorly for LLMs.
The null-sentence regime introduces severe distortions — most notably attention sink effects — that dominate the attribution signal.
Moreover, this degradation comes at a substantial computational cost: in practice, IG requires approximately $20\times$ more computation than Semantic Scope due to the discrete approximation of the path integral.
As a result, path integration both worsens attribution quality and significantly increases computational overhead in the settings explored in this work.

\subsection{Geometric interpretation of influence under \texorpdfstring{$\varepsilon$}{ep}
-norm perturbations}
\label{app:epsilon_perturbation}

We provide a precise justification for the geometric interpretation of the influence score defined in \cref{def:influence}.
Recall that $\bm{J}_t := \partial \bm{y} / \partial \bm{x}_t \in \mathbb{R}^{d_{\text{model}} \times d_{\text{model}}}$ denotes the Jacobian of the leading hidden state $\bm{y}$ with respect to the input embedding $\bm{x}_t$, and let $\bm{v} \in \mathbb{R}^{d_{\text{model}}}$ denote a fixed direction in output space.

Under a first-order Taylor expansion, a small perturbation $\bm{\delta x}_t$ to the input embedding induces a change in the output
\[
 \bm{\delta y} \approx\bm{J}_t \bm{\delta x}_t.
\]
Projecting this change onto the direction $\bm{v}$ yields the scalar response
\[
\bm{v}^\intercal \bm{\delta y}
\;\approx\;
\bm{v}^\intercal \bm{J}_t\bm{\delta x}_t.
\]

We are interested in the maximum possible change along $\bm{v}$ induced by any perturbation of bounded norm, i.e.,
\[
\max_{\|\bm{\delta x}_t\|_2 \le \varepsilon}
\bm{v}^\intercal \bm{J}_t \bm{\delta x}_t.
\]
By the Cauchy--Schwarz inequality, this quantity is upper-bounded by
\[
\bm{v}^\intercal \bm{J}_t \bm{\delta x}_t
\;\le\;
\|\bm{v}^\intercal \bm{J}_t\|_2 \, \|\bm{\delta x}_t\|_2
\;\le\;
\varepsilon \, \|\bm{v}^\intercal \bm{J}_t\|_2,
\]
with equality achieved when $\bm{\delta x}_t$ is aligned with $(\bm{v}^\intercal \bm{J}_t)^\intercal$.
Therefore,
\[
\max_{\|\bm{\delta x}_t\|_2 \le \varepsilon}
\bm{v}^\intercal \bm{J}_t \bm{\delta x}_t
=
\varepsilon \, \|\bm{v}^\intercal \bm{J}_t\|_2.
\]

This shows that $\|\bm{v}^\intercal \bm{J}_t\|_2$ quantifies the largest possible first-order magnification, along the direction $\bm{v}$, that can be induced by an $\varepsilon$-norm perturbation to the input token embedding $\bm{x}_t$, up to the scale factor $\varepsilon$.
This motivates the definition of the influence score in \cref{def:influence}.

\subsection{Second-order KL expansion and Fisher pullback}
\label{app:kl_second_order}

Let $\bm{p}(\cdot\mid \bm{x}_t) \in \mathbb{R}^{|\mathcal{V}|}$ denote a smooth family of
categorical distributions parameterized by the input~$\bm{x}_t$.
For a small perturbation $\bm{\delta x}_t$, consider the KL divergence
\begin{align}
&\mathrm{KL}\!\left(
\bm{p}(\cdot\mid \bm{x}_t)
\,\middle\|\,
\bm{p}(\cdot\mid \bm{x}_t+\bm{\delta x}_t)
\right) \notag \\
&=
\sum_{i}
\bm{p}(i\mid \bm{x}_t)
\Bigl(
\log \bm{p}(i\mid \bm{x}_t)
-
\log \bm{p}(i\mid \bm{x}_t+\bm{\delta x}_t)
\Bigr).
\notag
\end{align}
Under standard regularity conditions, the first-order term vanishes and
the KL divergence admits the second-order expansion
\begin{align}
\mathrm{KL}\!\left(
\bm{p}(\cdot\mid \bm{x}_t)
\,\middle\|\,
\bm{p}(\cdot\mid \bm{x}_t+\bm{\delta x}_t)
\right)
&=
\frac{1}{2}\,
\bm{\delta x}_t^\intercal
\bm{F}(\bm{x}_t)
\bm{\delta x}_t
+
o\!\left(\|\bm{\delta x}_t\|^2\right),
\notag
\end{align}
where the Fisher information matrix is given by \cite{amari2000methods}
\begin{align}
\bm{F}_t = \bm{F}(\bm{x}_t)
&:=
\mathbb{E}_{\bm{p}(\cdot\mid \bm{x}_t)}
\!\left[
(\nabla_{\bm{x}_t} \log \bm{p})
(\nabla_{\bm{x}_t} \log \bm{p})^\intercal
\right].
\notag
\end{align}
We now derive the closed-form expression for $\bm{F}_t$.
For a categorical distribution $\bm{p} = \mathrm{softmax}(\bm{z})$,
parameterized by logits $\bm{z} \in \mathbb{R}^{|\mathcal{V}|}$,
its Fisher information with respect to $\bm{z}$ takes the closed form
\cite{bishop2006pattern}
\begin{align}
\bm{F}(\bm{z})
&=
\mathrm{Diag}(\bm{p}) - \bm{p}\bm{p}^\intercal .
\notag
\end{align}

If the logits are given by $\bm{z} = \bm{W} \bm{y}$, then the Fisher
information with respect to the hidden state~$\bm{y}$ is
\begin{align}
\bm{F}:=\bm{F}(\bm{y})
&=
\bm{W}^\intercal
\bigl(
\mathrm{Diag}(\bm{p}) - \bm{p}\bm{p}^\intercal
\bigr)
\bm{W}.
\notag
\end{align}

Finally, to assess how an input token $\bm{x}_t$ influences the
model’s prediction, we pull back \cite{arvanitidis2022pullinginformationgeometry} the Fisher information matrix through the Jacobian $\bm{J}_t$ to the embedding space of token $t$:
\begin{equation}
\bm{F}_t = \bm{J}_t^\intercal \bm{F} \bm{J}_t  
\end{equation}

\noindent\textbf{Exact Hessian of the KL.}
The pullback $\bm{J}_t^\intercal \bm{F} \bm{J}_t$ is exact rather than a Gauss--Newton approximation.
Let $D(\bm{y}) := \mathrm{KL}\bigl(\bm{p}(\cdot\mid\bm{y}_0)\,\|\,\bm{p}(\cdot\mid\bm{y})\bigr)$, where $\bm{y}_0$ is the unperturbed hidden state and $\bm{y}$ is a function of $\bm{x}_t$.
By the chain rule,
\begin{equation}
\nabla^2_{\bm{x}_t} D
=
\bm{J}_t^\intercal \bigl(\nabla^2_{\bm{y}} D\bigr) \bm{J}_t
+
\sum_{k=1}^{d_{\mathrm{model}}} \frac{\partial D}{\partial y_k}\, \nabla^2_{\bm{x}_t} y_k .
\end{equation}
At the unperturbed input, $\nabla_{\bm{y}} D = \bm{0}$ because the KL divergence attains its minimum there, so the second term, which carries the curvature of the network itself, vanishes identically.
Moreover, $\nabla^2_{\bm{y}} D = \bm{W}^\intercal\bigl(\mathrm{Diag}(\bm{p}) - \bm{p}\bm{p}^\intercal\bigr)\bm{W} = \bm{F}$ at $\bm{y} = \bm{y}_0$.
Hence $\bm{J}_t^\intercal \bm{F} \bm{J}_t$ is the exact Hessian of the KL divergence with respect to $\bm{x}_t$ at $\bm{\delta x}_t = \bm{0}$, obtained using only the first-order Jacobian $\bm{J}_t$.

\subsection{Fisher Scope as a low-rank approximation of total information}
\label{app:fisher_total_information}

The Fisher Scope influence score in \cref{eq: fisher influence} is based on 
the principal Fisher direction $\bm{u}_1$, the single direction in hidden 
space most sensitive to changes in the predictive distribution. Here we show 
that this can be understood as the leading term of a more general quantity, which aggregates distributional sensitivity 
across all directions.

\noindent\textbf{Total information as expected KL divergence.}

Consider an isotropic perturbation $\bm{\delta x}_t= \varepsilon \bm{\xi}$ of norm $\varepsilon$, 
where $\varepsilon > 0$ is small and $\bm{\xi} \in \mathbb{R}^{d_{\mathrm{model}}}$ is uniform on 
the unit sphere. 
Taking the expectation and using
$\mathbb{E}_{\bm{\xi}}[\bm{\xi}\bm{\xi}^\intercal]=\tfrac{1}{d_{\mathrm{model}}}\bm{I}$ yields
\begin{align}
&\mathbb{E}_{\bm{\xi}}\!\left[
\mathrm{KL}\!\left(\bm{p}(\cdot\mid \bm{x}_t) \,\middle\|\, \bm{p}(\cdot\mid \bm{x}_t+\varepsilon\bm{\xi})\right)
\right] \notag \\
&=
\frac{1}{2}\,\varepsilon^2\,
\mathbb{E}_{\bm{\xi}}\!\left[\bm{\xi}^\intercal \bm{F}_t\,\bm{\xi}\right]
+ o(\varepsilon^2)
\notag\\
&=
\frac{1}{2}\,\varepsilon^2\,
\mathrm{tr}\!\Big(\bm{F}_t\,\mathbb{E}_{\bm{\xi}}[\bm{\xi}\bm{\xi}^\intercal]\Big)
+ o(\varepsilon^2)
\notag\\
&=
\frac{1}{2}\,\varepsilon^2\,
\frac{1}{d_{\mathrm{model}}}\,
\mathrm{tr}(\bm{F}_t)
+ o(\varepsilon^2).
\label{eq:expected_kl_trace}
\end{align}
Thus, up to the universal factor $\varepsilon^2/(2d_{\mathrm{model}})$, the trace 
$\mathrm{tr}(\bm{F}_t)$ governs the expected KL divergence induced by an 
isotropic perturbation of the input token $\bm{x}_t$. We therefore refer 
to $\mathrm{tr}(\bm{F}_t)$ as the \emph{total information} at position $t$.

\noindent\textbf{Fisher Scope as rank-1 approximation.}
Since $\bm{F}$ admits the eigen-decomposition $\bm{F} = \bm{U}\bm{\Lambda}\bm{U}^\intercal$, 
the total information can be written exactly as
\begin{equation}
    \mathrm{tr}(\bm{F}_t) = \sum_{i=1}^{d_{\mathrm{model}}} 
    \lambda_i \|\bm{u}_i^\intercal \bm{J}_t\|_2^2.
\end{equation}
In practice, $\bm{F}$ admits a good low-rank approximation 
$\bm{F} \approx \bm{U}_r \bm{\Lambda}_r \bm{U}_r^\intercal$, 
where $\bm{U}_r \in \mathbb{R}^{d_{\mathrm{model}} \times r}$ contains 
the top $r$ eigenvectors and $\bm{\Lambda}_r$ the corresponding eigenvalues. 
This yields the approximation
\begin{equation}
    \mathrm{tr}(\bm{F}_t) \approx \sum_{i=1}^{r} 
    \lambda_i \|\bm{u}_i^\intercal \bm{J}_t\|_2^2,
    \label{eq:low_rank_trace}
\end{equation}
where each term $\|\bm{u}_i^\intercal \bm{J}_t\|_2$ requires only a single 
backward pass, analogous to \cref{influence via auto-diff}. The full total 
information thus requires $r$ backward passes, compared to 
$d_{\mathrm{model}}$ passes needed to form $\bm{J}_t$ explicitly.

The Fisher Scope influence score in \cref{eq: fisher influence} corresponds 
to the $r=1$ case of \cref{eq:low_rank_trace}, retaining only the leading 
term $\lambda_1 \|\bm{u}_1^\intercal \bm{J}_t\|_2^2$ and dropping the 
scalar $\lambda_1$ since it is constant across token positions $t$.

\begin{figure}[h]
  \includegraphics[width=0.9\columnwidth]{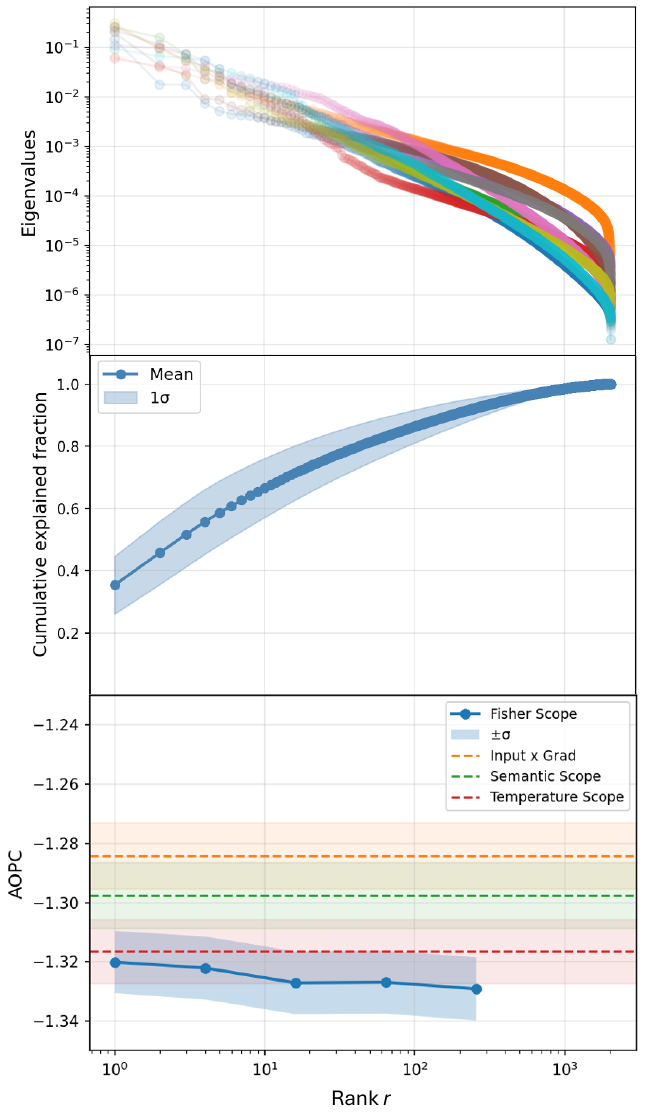}
  \caption{Justification for the rank-1 approximation used in Fisher Scope, evaluated on LLaMA-3.2 1B and LAMBADA.
  \textbf{Top:} Eigenvalue spectrum of $\bm{F}$ across 10 prompts on a log-log scale, exhibiting a consistent power-law decay spanning 6 orders of magnitude.
  \textbf{Middle:} Mean cumulative explained variance $\sum_{i=1}^r \lambda_i / \mathrm{tr}(\bm{F})$ averaged over 1000 passages (shaded band: $\pm 1\sigma$); $\bm{u}_1$ alone accounts for $\approx$37\% of total sensitivity (1$\sigma$: 25--50\%).
  \textbf{Bottom:} AOPC as a function of rank $r \in \{1, 4, 16, 64, 256\}$, with baselines shown as dashed lines (reproduced in \cref{fig:fisher_rank_aopc}). Rank-1 Fisher Scope already outperforms Input $\times$ Gradient and Semantic Scope; increasing $r$ yields only marginal improvement at $r$-fold computational cost.}
  \label{fig:fisher_spectrum}
\end{figure}

\noindent\textbf{Low-rankness of $\bm{F}$.}
\Cref{fig:fisher_spectrum} provides three complementary pieces of evidence that the $r=1$ approximation is justified in practice.
The top panel shows the eigenvalue spectrum of $\bm{F}$ across 10 LAMBADA prompts on a log-log scale: a consistent power-law decay spanning 6 orders of magnitude, indicating that the leading eigenvalue dominates all subsequent ones.

The middle panel confirms this quantitatively via the cumulative explained variance $\sum_{i=1}^r \lambda_i / \mathrm{tr}(\bm{F})$, averaged over 1000 LAMBADA passages: 
at $r=1$ the mean explained fraction is already $\approx 37\%$, rising to $\approx 65\%$ by $r=10$ and approaching 1 only beyond $r \approx 1000$.

The bottom panel (also \cref{fig:fisher_rank_aopc} in the main text) provides direct empirical justification, extending the LLaMA-3.2 1B results in \cref{tab:aupc}: rank-1 Fisher Scope ($r=1$) already outperforms Input $\times$ Gradient and Semantic Scope on LAMBADA, and matches Temperature Scope.
Increasing the rank to $r \in \{4, 16, 64, 256\}$ yields only marginal further gains in AOPC, while multiplying the computational cost by $r$.

Together, these results justify both the privileged status of $\bm{u}_1$ and the choice of $r=1$ in Fisher Scope.

\clearpage
\section*{NeurIPS Paper Checklist}

\begin{enumerate}

\item {\bf Claims}
    \item[] Question: Do the main claims made in the abstract and introduction accurately reflect the paper's contributions and scope?
    \item[] Answer: \answerYes{} 
    \item[] Justification: The abstract and \cref{sec:methodology} clearly state the paper's contributions: a Jacobian-based framework for input--output attribution in LLMs, three concrete instantiations (Semantic Scope, Fisher Scope, and Temperature Scope), and empirical evaluation on LAMBADA and IWSLT2017 benchmarks. Limitations and scope are discussed in the Limitations section.

\item {\bf Limitations}
    \item[] Question: Does the paper discuss the limitations of the work performed by the authors?
    \item[] Answer: \answerYes{} 
    \item[] Justification: The paper includes a dedicated Limitations section discussing these points.

\item {\bf Theory assumptions and proofs}
    \item[] Question: For each theoretical result, does the paper provide the full set of assumptions and a complete (and correct) proof?
    \item[] Answer: \answerYes{} 
    \item[] Justification: \cref{app:epsilon_perturbation,app:kl_second_order,app:fisher_total_information}


    \item {\bf Experimental result reproducibility}
    \item[] Question: Does the paper fully disclose all the information needed to reproduce the main experimental results of the paper to the extent that it affects the main claims and/or conclusions of the paper (regardless of whether the code and data are provided or not)?
    \item[] Answer: \answerYes{} 
    \item[] Justification: \cref{sec:methodology,sec:evaluation} provide sufficient detail to reproduce all results.

\item {\bf Open access to data and code}
    \item[] Question: Does the paper provide open access to the data and code, with sufficient instructions to faithfully reproduce the main experimental results, as described in supplemental material?
    \item[] Answer: \answerYes{} 
    \item[] Justification: Code and data will be released upon acceptance. The paper provides sufficient methodological detail to reproduce the core methods: the three influence score formulations are given in closed form (Equations 4, 6, 8), the benchmark datasets (LAMBADA and IWSLT2017) are publicly available, the base models (LLaMA-3.2, Qwen2.5, Gemma-3) are open-source, and the evaluation metric (AOPC) is described in full in Section 5 and Appendix A.1.

\item {\bf Experimental setting/details}
    \item[] Question: Does the paper specify all the training and test details (e.g., data splits, hyperparameters, how they were chosen, type of optimizer) necessary to understand the results?
    \item[] Answer: \answerYes{} 
    \item[] Justification: Details regarding dataset and benchmark setup can be found in \cref{sec:evaluation,sec:benchmark_details}.

\item {\bf Experiment statistical significance}
    \item[] Question: Does the paper report error bars suitably and correctly defined or other appropriate information about the statistical significance of the experiments?
    \item[] Answer: \answerYes{} 
    \item[] Justification: We provide error bars in \cref{sec:evaluation}

\item {\bf Experiments compute resources}
    \item[] Question: For each experiment, does the paper provide sufficient information on the computer resources (type of compute workers, memory, time of execution) needed to reproduce the experiments?
    \item[] Answer: \answerYes{} 
    \item[] Justification: We report the GPU and clock time in the Limitations section.
    
\item {\bf Code of ethics}
    \item[] Question: Does the research conducted in the paper conform, in every respect, with the NeurIPS Code of Ethics \url{https://neurips.cc/public/EthicsGuidelines}?
    \item[] Answer: \answerYes{} 
    \item[] Justification: We follow the NeurIPS Code of Ethics.

\item {\bf Broader impacts}
    \item[] Question: Does the paper discuss both potential positive societal impacts and negative societal impacts of the work performed?
    \item[] Answer: \answerYes{} 
    \item[] Justification: \cref{sec:Semantic Scope} and \cref{fig:causal attribution 2} in the main paper demonstrate how Semantic Scope can surface implicit political biases in LLMs, illustrating a direct application to model auditing and fairness analysis. We regard the negative risk as relatively indirect: Jacobian Scopes are a diagnostic tool that characterizes, but does not modify, model behaviour, and confer no additional control over the model itself.
    
\item {\bf Safeguards}
    \item[] Question: Does the paper describe safeguards that have been put in place for responsible release of data or models that have a high risk for misuse (e.g., pre-trained language models, image generators, or scraped datasets)?
    \item[] Answer: \answerNA{} 
    \item[] Justification: Our work does not pose such risks.

\item {\bf Licenses for existing assets}
    \item[] Question: Are the creators or original owners of assets (e.g., code, data, models), used in the paper, properly credited and are the license and terms of use explicitly mentioned and properly respected?
    \item[] Answer: \answerYes{} 
    \item[] Justification: We cite the creators of the LLMs and datasets used throughout the paper.

\item {\bf New assets}
    \item[] Question: Are new assets introduced in the paper well documented and is the documentation provided alongside the assets?
    \item[] Answer: \answerNA{} 
    \item[] Justification: No new assets are introduced.

\item {\bf Crowdsourcing and research with human subjects}
    \item[] Question: For crowdsourcing experiments and research with human subjects, does the paper include the full text of instructions given to participants and screenshots, if applicable, as well as details about compensation (if any)? 
    \item[] Answer: \answerNA{} 
    \item[] Justification: Our work does not involve crowdsourcing or research with human subjects.

\item {\bf Institutional review board (IRB) approvals or equivalent for research with human subjects}
    \item[] Question: Does the paper describe potential risks incurred by study participants, whether such risks were disclosed to the subjects, and whether Institutional Review Board (IRB) approvals (or an equivalent approval/review based on the requirements of your country or institution) were obtained?
    \item[] Answer: \answerNA{} 
    \item[] Justification: Our work does not involve crowdsourcing or research with human subjects.

\item {\bf Declaration of LLM usage}
    \item[] Question: Does the paper describe the usage of LLMs if it is an important, original, or non-standard component of the core methods in this research? Note that if the LLM is used only for writing, editing, or formatting purposes and does \emph{not} impact the core methodology, scientific rigor, or originality of the research, declaration is not required.
    \item[] Answer: \answerNA{} 
    \item[] Justification: The core method development in our work does not involve LLMs as any important, original, or non-standard components.

\end{enumerate}
\end{document}